%% file: main.tex
\documentclass[10pt,twocolumn,a4paper]{article}
\usepackage{cite}
\usepackage[margin=0.77in,columnsep=0.25in]{geometry}
\usepackage{amsmath,amssymb,amsfonts,bm}
\usepackage{algorithmic}
\usepackage{graphicx}
\usepackage{textcomp}
\usepackage{booktabs}
\usepackage{multirow}
\usepackage{threeparttable}
\usepackage[table]{xcolor}
\definecolor{mimicfour}{HTML}{F0F4F8}
\definecolor{mimicthree}{HTML}{F9EBEA}
\usepackage{colortbl}
\usepackage{adjustbox}
\usepackage{microtype}
\usepackage{hyperref}
\usepackage{subcaption}
\usepackage[figuresright]{rotating}
\usepackage{svg}
\hypersetup{hidelinks=true}

\def\BibTeX{{\rm B\kern-.05em{\sc i\kern-.025em b}\kern-.08em
    T\kern-.1667em\lower.7ex\hbox{E}\kern-.125emX}}

\newcommand{\modelname}{\textsc{TRACE}}

\newcommand{\gain}[1]{\textbf{+#1}}
\newcommand{\dropval}[1]{$-$#1}

\begin{document}

\title{\modelname{}: Tracking Multimodal Clinical Trajectories for Inpatient Risk Prediction}

\author{
    Minjun Kim\thanks{Equal contribution.} 
    \and 
    Jong Hak Moon\textsuperscript{*}\thanks{Corresponding author: Jong Hak Moon. Both authors are with Yeji X (\texttt{mj.kim@yejix.ai}; \texttt{jh.moon@yejix.ai}).}
}
\date{}

\maketitle

\input{0.abs}

\section{Introduction}
\label{sec:introduction}
\input{1.intro}

\section{Methods}
\label{sec:methods}
\input{2.methods}

\section{Experiments}
\label{sec:experiments}
\input{3.exp}

\section{Discussion}
\label{sec:discussion}
\input{discussion}

\section{Conclusion}
\label{sec:conclusion}
\input{4.conclusion}

\bibliographystyle{IEEEtran}
\bibliography{ref}

\end{document}

%% file: 0.abs.tex
\begin{abstract}
Accurate inpatient risk prediction requires modeling heterogeneous clinical evidence that accumulates irregularly over time. This challenge spans in-hospital mortality, prolonged length of stay, and readmission, each with task-specific observation windows. Existing multimodal approaches combine structured health record data, chest x-rays, and clinical text, but primarily focus on modality fusion, admission-level relational structure, or missing-modality compensation, with limited attention to how clinical evidence evolves as a temporal trajectory over the course of an admission. We propose \modelname{} (\textbf{TRA}jectory modeling of multimodal \textbf{C}linical \textbf{E}vents), a daily-scale global--local framework that organizes recorded clinical events into a day-level trajectory, separating day-invariant global context, such as demographics and diagnosis-derived information, from day-specific local events drawn from structured EHR data, chest x-rays, radiology reports, and discharge summaries. The resulting trajectory is encoded using temporal global--local attention, allowing unavailable modalities to emit no tokens rather than being imputed. A phenotype-guided contrastive objective further organizes admissions according to both outcome status and diagnosis-derived phenotype, providing additional structure under heterogeneous disease distributions. We construct cohorts from MIMIC-IV, eICU, and MIMIC-III comprising nearly 180,000 admissions and ICU stays, and evaluate \modelname{} on in-hospital mortality, prolonged length of stay, and 30-day readmission prediction. \modelname{} achieves the highest AUROC across all nine dataset--task settings and the highest ACC in eight, compared with fusion-centered, relational, and missing-modality-aware baselines. Ablation studies show that the temporal global--local attention mask achieves the highest AUROC across all three tasks on MIMIC-IV, while phenotype-guided contrastive learning improves mean phenotype-stratified AUROC across all nine dataset--task settings. Trajectory-level attribution further localizes the hospital days and modalities associated with changes in predicted risk. 
\end{abstract}

\par\noindent\textbf{Keywords:} 
Inpatient risk prediction, multimodal clinical trajectory modeling, electronic health records, temporal attention, supervised contrastive learning

%% file: 1.intro.tex
\begin{figure*}[t]
  \centering
    \includegraphics[height=0.33\textheight, width=\textwidth]{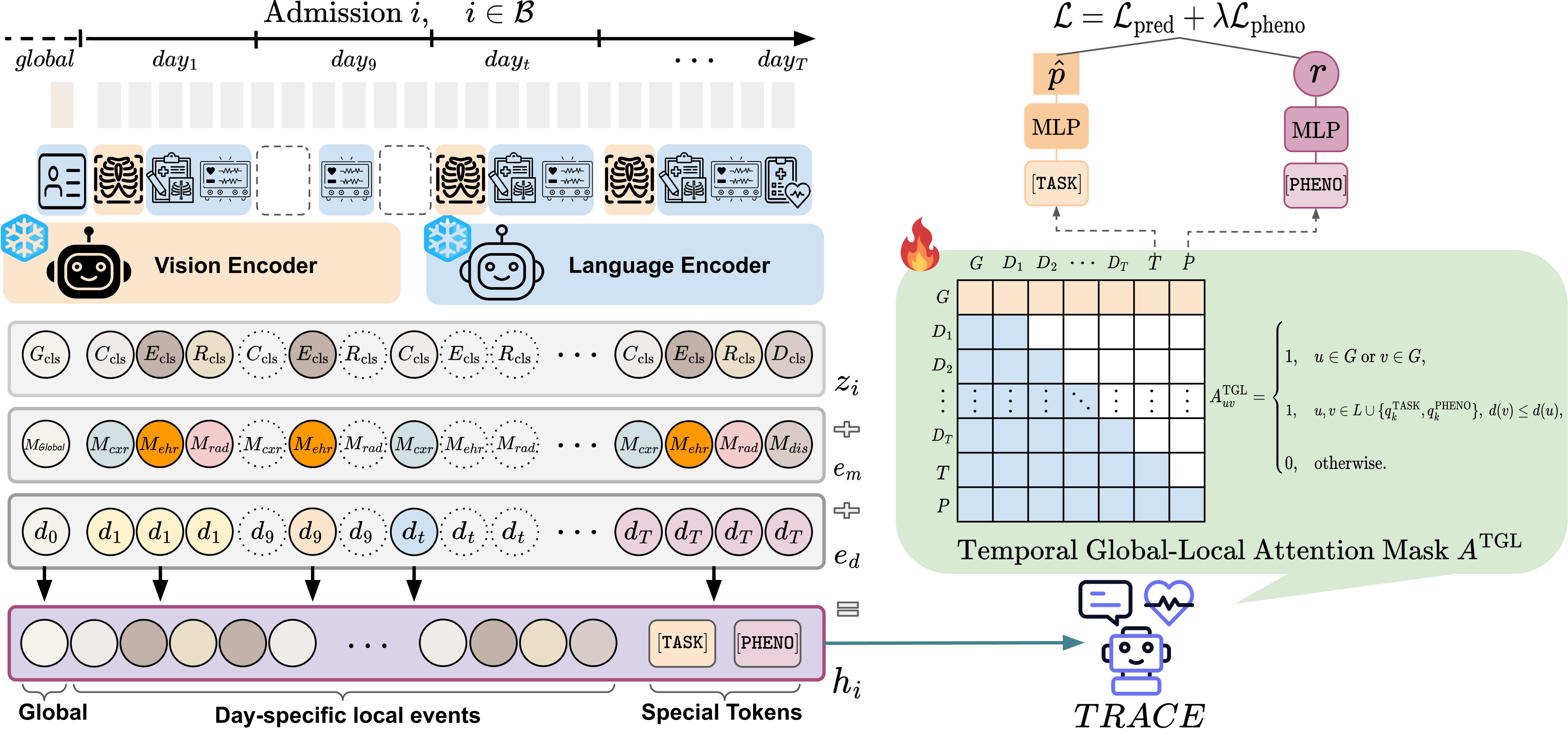}
\caption{\footnotesize \textbf{Overview of \modelname{}.} Admission-level global categories and local inputs recorded within the task-specific observation window are encoded using frozen pretrained vision or language encoders, projected into a shared space, and augmented with modality and day embeddings. For visual brevity, the category-specific global tokens $\mathbf{G}_i$ are shown collectively as $G_{\mathrm{cls}}$. Global tokens, day-specific local tokens $\mathbf{L}_i^{(k)}$, and learnable \texttt{[TASK]} and \texttt{[PHENO]} tokens are encoded by a masked Transformer using the temporal global--local attention mask $\mathbf{A}^{\mathrm{TGL}}$, which keeps global context accessible, preserves temporal order among local events, and allows \texttt{[PHENO]} to attend to \texttt{[TASK]} but not the reverse. The encoded \texttt{[TASK]} and \texttt{[PHENO]} states produce $\hat{p}^{(k)}$ and $\mathbf{r}^{(k)}$, respectively, and are jointly optimized using $\mathcal{L}^{(k)}=\mathcal{L}_{\mathrm{pred}}^{(k)} +\lambda\mathcal{L}_{\mathrm{pheno}}^{(k)}$. Snowflake and flame icons indicate frozen and trainable modules.}  
\label{fig:overview}
\end{figure*}

Inpatient risk prediction is challenging because patient status evolves over time and is documented through heterogeneous, irregularly recorded clinical events~\cite{lyu2023multimodal, multimodalpred, harutyunyan2019multitask, xie2022deep, morid2023time, crl-mmnar, muse, m3care, adacomed, mmstgnn, must}. During hospitalization, clinicians integrate relatively stable patient context, such as demographics and diagnosis-derived information, with evolving signals, including laboratory measurements, medications, procedures, imaging findings, and clinical notes~\cite{mmstgnn, m3care, adacomed, crl-mmnar, muse}. These data do not form a complete or uniformly sampled multimodal record. Instead, each admission contains a patient- and dataset-specific subset of modalities, and the meaning of the recorded events depends not only on what is observed, but also on when it occurs and how it changes over time~\cite{crl-mmnar, muse, m3care}.

Inpatient risk prediction is commonly formulated as admission-level binary classification, in which the model estimates the probability of an adverse outcome, such as prolonged length of stay (Long LOS), in-hospital mortality, and 30-day readmission~\cite{lyu2023multimodal, multimodalpred, crl-mmnar, muse, m3care, adacomed, mmstgnn, must}.
The clinically appropriate observation window differs across these binary prediction tasks. Long LOS prediction is most useful early in the admission for resource planning~\cite{stone2022systematic, el2021machine, soenksen2022integrated}, whereas in-hospital mortality prediction and 30-day readmission prediction may use data recorded over the completed hospitalization~\cite{crl-mmnar,adacomed,muse,m3care,mmstgnn, multimodalpred}. 
A unified trajectory model should therefore support task-specific observation windows while preserving the temporal organization of the clinical record.

The central modeling challenge lies in how the admission representation is
constructed within each observation window. Compressing the available data into a single static representation can obscure day-specific changes in patient status~\cite{xie2022deep, morid2023time}, while conventional multimodal fusion often assumes a fixed input structure that does not reflect the incomplete and irregular nature of real-world clinical records~\cite{zhou2023transformer, van2024yet, adacomed, crl-mmnar, muse, must, m3care}. A rigid organization of inputs may also fail to distinguish early prediction settings from completed-admission assessment.

Prior clinical AI models have addressed important aspects of this problem. Multimodal fusion methods combine structured EHR, imaging, and clinical text~\cite{adacomed, crl-mmnar, m3care, muse, multimodalpred, zhou2023transformer}; graph-based approaches~\cite{mmstgnn} incorporate admission similarity or relational structure; and missing-modality methods~\cite{crl-mmnar, m3care, muse} improve learning when some modalities are unavailable. These directions have substantially advanced admission-level clinical prediction, but with limited attention to how clinical evidence evolves as a temporal trajectory over the course of an admission. Less directly addressed is how to construct a unified admission representation that reflects this trajectory at the day level, separates stable global context from evolving local events, and accommodates task-specific temporal windows across datasets with different modality availability.

We propose \modelname{} (\textbf{TRA}jectory modeling of multimodal \textbf{C}linical \textbf{E}vents), a daily-scale global--local framework for multimodal inpatient risk prediction. \modelname{} retains the admission-level prediction objective but changes how the clinical record is represented before the final task-specific prediction. For each task, the model organizes multimodal clinical events recorded within the corresponding observation window along the admission timeline, separating day-invariant global context, such as demographic, diagnosis-derived, or severity-related information, from day-specific local events, including structured EHR events, imaging studies, radiology reports, and clinical documentation. Because each local event is tied to a specific day and modality, real-world irregularity in modality availability is handled naturally: rather than forcing each admission to conform to a complete multimodal grid, \modelname{} operates directly on whatever is actually recorded, with missing modalities simply emitting no tokens. This event-driven representation preserves modality identity, hospital-day position, and the temporal structure of the recorded clinical data.

To encode the resulting trajectory, \modelname{} uses temporal global--local (TGL) attention. Global context tokens remain accessible across the trajectory because they are not tied to a specific hospital day, whereas local event tokens follow the temporal order of the admission and support same-day multimodal fusion. This design combines stable patient context with evolving daily events without reconstructing unobserved modalities.

\modelname{} further incorporates phenotype-guided supervised contrastive learning to organize admission representations. Patients reaching the same endpoint may follow heterogeneous disease trajectories~\cite{placido2023deep, oikonomou2023machine}, and prior work has demonstrated the value of contrastive learning for structuring clinical representations~\cite{wanyan2021contrastive}. We therefore define positive pairs using both outcome status and diagnosis-derived phenotype~\cite{vu2021medaug, hager2023best}.

We evaluate \modelname{} on MIMIC-IV~\cite{PhysioNet-mimiciv-3.1}, eICU~\cite{eicu}, and MIMIC-III~\cite{mimic-iii} across three clinically distinct binary prediction settings: in-hospital mortality prediction using the completed admission, early Long LOS prediction from the first 48 hours, and discharge-time prediction of 30-day readmission. Across these datasets and tasks, \modelname{} achieves consistent discrimination gains over fusion-centered, relational, and missing-modality-aware baselines. Ablation studies evaluate the contribution of individual modalities, temporal global--local attention, and phenotype-guided contrastive learning. Finally, trajectory-level attribution localizes hospital days and modality-specific inputs associated with changes in predicted risk.

Our contributions are summarized as follows:
\begin{itemize}
    \item We propose \modelname{}, a daily-scale multimodal framework that separates day-invariant global context from day-specific local events and encodes them as a temporal global--local trajectory using temporal global--local (TGL) attention within each task-specific observation window.
    \item We incorporate phenotype-guided contrastive learning to structure admission representations jointly by outcome status and diagnosis-derived phenotype.
    \item We evaluate \modelname{} on MIMIC-IV, eICU, and MIMIC-III across in-hospital mortality, Long LOS, and 30-day readmission prediction, achieving the highest AUROC across all nine dataset--task settings.
    \item We show that trajectory-level attribution localizes the hospital days and modalities driving changes in predicted risk, offering an interpretable view of how predictions evolve over an admission.
\end{itemize}

\section{Related Work}

\subsection{Multimodal Fusion and Relational Clinical Risk Prediction}

Clinical risk prediction has increasingly expanded from unimodal structured EHR modeling to multimodal approaches that combine structured EHR data, medical images, and clinical text~\cite{miotto2018deep, adacomed,mmstgnn, crl-mmnar,m3care,muse, must, multimodalpred}. MultimodalPred~\cite{multimodalpred} combines temporal variables, chest x-rays, and clinical notes for in-hospital mortality, Long LOS, and readmission prediction. 
Chen et al.~\cite{adacomed} report both the Large-Ensemble baseline and AdaCoMed, which integrate modality-specific predictors through ensemble-based and collaborative strategies, respectively. Graph-based models such as MM-STGNN~\cite{mmstgnn} additionally incorporate longitudinal EHR and imaging data with admission-similarity structure. These studies establish the complementary value of multimodal and temporal
clinical data. However, temporal information is often modeled within
individual modalities, aggregated at the admission level, or used for
admission-level relational propagation. In contrast, \modelname{} places the
admission trajectory itself at the center of the representation by organizing
recorded multimodal events according to hospital day while preserving modality identity and temporal position. 

\subsection{Missing-Modality-Aware Clinical Representation Learning}

Real-world multimodal clinical records are incomplete and irregularly
observed, with modality availability varying across patients, hospital days, and datasets. M3Care~\cite{m3care} estimates task-relevant missing representations using similar patients, whereas MUSE/MUSE+~\cite{muse} model patient--modality relationships through bipartite graphs and contrastive learning. CRL-MMNAR~\cite{crl-mmnar} further accounts for clinically informative, non-random modality missingness. These methods primarily reconstruct, compensate for, or exploit unavailable
modalities. \modelname{} addresses a complementary problem: organizing the clinical events that are actually recorded without reconstructing absent
modalities. An unavailable modality contributes no token, while each recorded
event remains aligned with its hospital day. 
In addition, rather than using
contrastive learning for cross-modal alignment or missing-modality recovery,
\modelname{} defines positive pairs using both outcome labels and
diagnosis-derived phenotypes, encouraging clinically meaningful organization of
the admission representation.

%% file: 2.methods.tex
This section formalizes the \modelname{} architecture introduced in Section~\ref{sec:introduction}. Section~\ref{sec:problem} defines the task-specific observation window and prediction objective; Section~\ref{sec:trajectory_rep} describes how multimodal clinical events are converted into an ordered token sequence; Section~\ref{sec:TGL} introduces the temporal global--local (TGL) attention mask used to encode this sequence; and Section~\ref{sec:objectives} defines the training objectives, including the phenotype-guided contrastive term.

\subsection{Problem Formulation}
\label{sec:problem}

We formulate three inpatient risk prediction tasks as binary
classification problems with task-specific observation windows:
in-hospital mortality, Long LOS, and 30-day readmission. For
admission $i$, let $T_i$ denote the total number of hospital days
and $H_i^{(k)}$ denote the final observation day for task $k$:
\[
H_i^{(k)} =
\begin{cases}
2, & k = \mathrm{LongLOS}, \\
T_i, & k \in \{\mathrm{Mortality}, \mathrm{Readmission}\}.
\end{cases}
\]

The task-specific clinical record is represented as
\[
\mathcal{X}_i^{(k)} = \left(\mathcal{G}_i, \mathcal{E}_{i,1}, \ldots, \mathcal{E}_{i,H_i^{(k)}}\right),
\]
where $\mathcal{G}_i$ denotes day-invariant global patient context and
$\mathcal{E}_{i,d}$ denotes the set of multimodal clinical events and
documents recorded on hospital day $d$. Thus, each admission is
represented by global context together with longitudinal local
events and documents recorded within the observation window of the
corresponding task. For task $k$ with binary outcome label $p_i^{(k)} \in \{0, 1\}$, the
objective is to estimate
\[
\hat{p}_i^{(k)} = f_k\left(\mathcal{X}_i^{(k)}\right), \qquad \hat{p}_i^{(k)} \in [0, 1].
\]

Each task is trained independently using the same \modelname{}
architecture with a task-specific token and prediction head. Although
the outcome is defined at the admission level, prediction is based on
a temporally organized multimodal trajectory rather than a static
aggregation of the clinical record; concrete task definitions and
observation windows for each dataset are given in
Section~\ref{sec:datasets}.

\subsection{Multimodal Trajectory Representation}
\label{sec:trajectory_rep}

We transform $\mathcal{X}_i^{(k)}$ into an ordered token sequence
that preserves modality identity and hospital-day position. Let
$\mathcal{M}_{i,d}$ denote the set of modalities recorded for
admission $i$ on day $d$, drawn from $m \in \{\mathrm{C, E, R, D}\}$,
corresponding to chest x-rays, structured EHR events, the associated
chest x-ray radiology reports, and discharge summaries, respectively.
A modality that is unavailable in a dataset or not recorded on a
given day contributes no local token.

EHR-derived variables are assigned according to their temporal role. Admission-level information, such as demographics, diagnosis-derived context, or severity-related variables (e.g., APACHE), is represented as global context~\cite{mmstgnn, muse}. Time-varying variables, including laboratory measurements, medications, and treatments, are assigned to the corresponding hospital day as local clinical events~\cite{mmstgnn, m3care, crl-mmnar, adacomed, muse}.
Regardless of temporal role, these variables are converted into value-preserving natural-language or code-based descriptors that retain both the clinical concept and its recorded value.
All structured EHR events recorded on the same hospital day are concatenated into a single descriptor and encoded into one $E_{\mathrm{cls}}$ token per day via the clinical language encoder. Each chest x-ray image similarly yields one $C_{\mathrm{cls}}$ token via the vision encoder, while its
associated radiology report and the discharge summary each yield one $R_{\mathrm{cls}}$ or $D_{\mathrm{cls}}$ token, respectively, via the clinical language encoder; multiple images or reports recorded on the same day are each retained as a separate token rather than being merged.

For each observed modality $m \in \mathcal{M}_{i,d}$, let
$N_{i,d}^m$ denote the number of recorded items of modality $m$ on
day $d$, and let $x_{i,d,j}^m$ denote the $j$-th such item ($1 \le j
\le N_{i,d}^m$). For structured EHR events, $N_{i,d}^m = 1$ because
all events recorded on a given day are concatenated into a single
descriptor; for radiology reports, discharge summaries, and chest
x-rays, $N_{i,d}^m$ equals the number of available documents or
images on day $d$. Each item is encoded and projected into the
shared model space, yielding the modality-specific token shown in
Figure~\ref{fig:overview} (i.e., $C_{\mathrm{cls}}$,
$E_{\mathrm{cls}}$, $R_{\mathrm{cls}}$, or $D_{\mathrm{cls}}$):
\[
\mathbf{z}_{i,d,j}^m = \mathbf{W}_{\mathrm{mod}}^m E_{\tau(m)}\left(x_{i,d,j}^m\right) + \mathbf{b}_{\mathrm{mod}}^m, \qquad \mathbf{z}_{i,d,j}^m \in \mathbb{R}^{d_{\mathrm{model}}},
\]
where $E_{\tau(m)}$ denotes the encoder associated with modality
$m$, and $d_{\mathrm{model}}$ is the shared hidden dimension of
\modelname{}. Chest x-rays are encoded using a medical vision
encoder~\cite{eva-x}, whereas descriptor-based EHR inputs,
radiology reports, and discharge summaries are encoded using a
clinical language encoder~\cite{modernbert}. Text-based
modalities may share the same encoder while retaining
modality-specific projection layers and embeddings.

Each projected representation is augmented with a learnable modality
embedding $\mathbf{e}_m \in \{\mathbf{e}_{\mathrm{cxr}},
\mathbf{e}_{\mathrm{ehr}}, \mathbf{e}_{\mathrm{rad}},
\mathbf{e}_{\mathrm{dis}}\}$ (denoted $M_{\mathrm{cxr}}$,
$M_{\mathrm{ehr}}$, $M_{\mathrm{rad}}$, and $M_{\mathrm{dis}}$ in
Figure~\ref{fig:overview}) and a learnable day embedding
$\mathbf{e}_d \in \{\mathbf{e}_1, \ldots, \mathbf{e}_{T_i}\}$
(denoted $d_1, \ldots, d_T$ in Figure~\ref{fig:overview}):
\[
\mathbf{h}_{i,d,j}^m = \mathbf{z}_{i,d,j}^m + \mathbf{e}_m + \mathbf{e}_d, \qquad \mathbf{h}_{i,d,j}^m \in \mathbb{R}^{d_{\mathrm{model}}}.
\]
Multiple items of the same modality recorded on the same day share
the same modality and day embeddings but originate from distinct
inputs. The day embedding is indexed by the day within the
admission, up to the maximum sequence window specified in the
experimental setup.

Within each day, modality-specific tokens are serialized using a
fixed canonical order: $C_{\mathrm{cls}}$, $E_{\mathrm{cls}}$,
$R_{\mathrm{cls}}$, $D_{\mathrm{cls}}$; multiple items of the same
modality on the same day are ordered by document or acquisition
timestamp. Modalities that are unavailable or not recorded are
skipped.

Let $${
\mathbf{L}_i^{(k)} = \left\{ \mathbf{h}_{i,d,j}^m : 1 \le d \le H_i^{(k)},\; m \in \mathcal{M}_{i,d},\; 1 \le j \le N_{i,d}^m \right\}
}$$ denote the resulting ordered sequence of local modality tokens. The
global context $\mathcal{G}_i$ is constructed at the category level
rather than as a single pooled representation: each demographic
field (e.g., age, sex), the diagnosis-derived admission summary, and,
for eICU, the APACHE severity summary are each encoded independently
into their own token, so that $\mathbf{G}_i \in \mathbb{R}^{n_G
\times d_{\mathrm{model}}}$ consists of $n_G$ category-specific
tokens. For visual brevity, Figure~\ref{fig:overview} depicts this
set of category-specific tokens collectively as a single
$G_{\mathrm{cls}}$ block rather than enumerating each field
separately. Every global context token is assigned the shared global
modality embedding $\mathbf{e}_{\mathrm{global}}$ (denoted
$M_{\mathrm{Global}}$ in Figure~\ref{fig:overview}), marking it as
global rather than tied to a specific local modality, and the
reserved day embedding $\mathbf{e}_{d_0}$ (denoted $d_0$ in
Figure~\ref{fig:overview}), indicating that it is not tied to a
specific hospital day. The exact set of categories and the resulting
$n_G$ depend on dataset availability and are described in
Section~\ref{sec:datasets}. The task-specific input trajectory is
\[
\mathcal{H}_i^{(k)} = \left[\mathbf{G}_i; \mathbf{L}_i^{(k)}\right].
\]

Because unavailable modalities contribute no clinical tokens, trajectory length varies with the number of recorded events across admissions and datasets rather than conforming to a fixed multimodal input structure. Hospital-day granularity aligns with the natural cadence of clinical documentation, such as daily laboratory panels, physician notes, and imaging orders, while providing a practical trade-off among fine-grained event timing, data sparsity, and computational cost.

\subsection{Temporal Global--Local Trajectory Encoding}
\label{sec:TGL}

We encode the trajectory using a masked Transformer encoder with a
temporal global--local (TGL) attention mask. Given
$\mathcal{H}_i^{(k)}$, \modelname{} appends two learnable special
tokens to the observation-driven trajectory: a task token
$q_k^{\mathrm{TASK}}$ used for outcome prediction and a phenotype
token $q_k^{\mathrm{PHENO}}$ used for the contrastive objective,
\[
\mathcal{S}_i^{(k)} = \left[\mathbf{G}_i; \mathbf{L}_i^{(k)}; q_k^{\mathrm{TASK}}; q_k^{\mathrm{PHENO}}\right].
\]
Let $G$ denote the set of global context tokens and $L$ denote the
set of local event tokens. The two special tokens are treated
separately, $q_k^{\mathrm{TASK}}, q_k^{\mathrm{PHENO}} \notin G \cup
L$. Global context tokens represent day-invariant patient
information. For each local token $u \in L$, $d(u)$ denotes its
hospital day. We extend this temporal index to the two special tokens
by defining
\[
d\!\left(q_k^{\mathrm{TASK}}\right)=H_i^{(k)}+1,
\qquad
d\!\left(q_k^{\mathrm{PHENO}}\right)=H_i^{(k)}+2.
\]
For tokens $u$ and $v$ in $\mathcal{S}_i^{(k)}$, let $A_{uv} = 1$
indicate that query token $u$ may attend to key/value token $v$. The
TGL attention mask is defined as
\[
A_{uv}^{\mathrm{TGL}} =
\begin{cases}
1, & u \in G \text{ or } v \in G, \\
1, & u, v \in L \cup
\{q_k^{\mathrm{TASK}}, q_k^{\mathrm{PHENO}}\},\;
d(v) \le d(u), \\
0, & \text{otherwise}.
\end{cases}
\]
The phenotype token is appended after the task token in $\mathbf{S}_i^{(k)}$.
As a result, it attends to the complete observed trajectory, including the
task token itself. The task token, in turn, attends to all global and local
context but not to the phenotype token. This reflects a causal ordering
between the two special tokens: both tokens summarize the full global and
local context for their respective objectives. Global context tokens are
day-invariant only at input construction; through self-attention, their
encoded representations may incorporate information from local events.
Local tokens attend to global context and to events from the same or
earlier hospital days, while same-day tokens interact bidirectionally to
enable multimodal fusion without imposing an arbitrary within-day order.

After applying the masked Transformer encoder, the encoded sequence is denoted by $\tilde{\mathcal{S}}_i^{(k)}$. The hidden states at the task and phenotype token positions define the task-specific admission representation and the phenotype-guided representation, respectively:
\[
h_i^{(k)} = \tilde{q}_k^{\mathrm{TASK}}, \qquad h_i^{(k),\mathrm{pheno}} = \tilde{q}_k^{\mathrm{PHENO}}.
\]

\subsection{Training Objectives}
\label{sec:objectives}

\subsubsection{Admission-Level Risk Prediction}
The task-specific prediction head maps the admission representation
$h_i^{(k)}$ to the outcome probability:
\[
\hat{p}_i^{(k)} = \sigma\left(w_k^\top h_i^{(k)} + b_k\right),
\]
where $\sigma(\cdot)$ denotes the sigmoid function, and $w_k$ and
$b_k$ are task-specific parameters. The binary cross-entropy
prediction objective is
\[
\mathcal{L}_{\mathrm{pred}}^{(k)} = -\frac{1}{|\mathcal{B}|}\sum_{i\in\mathcal{B}}\left[p_i^{(k)}\log\hat{p}_i^{(k)} + \left(1-p_i^{(k)}\right)\log\left(1-\hat{p}_i^{(k)}\right)\right],
\]
where $\mathcal{B}$ denotes a training mini-batch.

\subsubsection{Phenotype-Guided Contrastive Objective}
The binary outcome label identifies the predicted risk state but
does not capture the heterogeneous clinical pathways through which
that outcome may arise. For example, two readmitted patients may
have circulatory and respiratory conditions, respectively, and may
not be clinically equivalent despite sharing the same outcome.
Conversely, patients with the same phenotype may have different
outcome labels. We therefore hypothesize that jointly conditioning
on outcome status and diagnosis-derived phenotype yields more
clinically meaningful positive pairs than outcome status alone, and
design the contrastive objective accordingly.

Each admission is assigned a diagnosis-derived phenotype group
$\phi_i \in \{1,\ldots,C_D\}$, where $C_D$ denotes the number of
phenotype groups in dataset $D$. Dataset-specific diagnosis codes are mapped to chapter-level clinical super-groups following the official ICD-9/10 classification.\footnote{\url{https://icd.who.int}}
When diagnoses from multiple phenotype groups were present, the
phenotype corresponding to the primary diagnosis was assigned as the single major phenotype. We use 14 phenotype groups for MIMIC-III and MIMIC-IV and 12 for eICU.\footnote{The 14 MIMIC phenotype groups are Circulatory, Respiratory, Nervous System, Digestive, Genitourinary, Endocrine \& Metabolic, Infectious, Blood \& Immune, Mental Disorders, Musculoskeletal \& Skin, Neoplasms, Symptoms \& Signs, Trauma \& Poisoning, and Factors \& Services. eICU uses the same groups except Mental Disorders and Musculoskeletal \& Skin.}

The phenotype-token representation is passed through an MLP projection head:
\[
\mathbf{r}_i^{(k)}
=
g_k\left(\mathbf{h}_i^{(k),\mathrm{pheno}}\right).
\]

For anchor admission $i$, the positive set is defined as
\[
\mathcal{P}_k(i)
=
\left\{
j \in \mathcal{B}\setminus\{i\}
\;\middle|\;
p_j^{(k)}=p_i^{(k)},\;
\phi_j=\phi_i
\right\}.
\]
Thus, two admissions form a positive pair only when they share both
the task label and phenotype group. Admissions that differ in either
attribute remain contrastive candidates but are not treated as
positives. Mini-batches are constructed using label- and
phenotype-aware sampling so that every anchor has at least one
positive example. Specifically, $(\phi,p^{(k)})$ groups are sampled
approximately in proportion to their prevalence, with at least two
admissions drawn from each selected group.

The phenotype-guided contrastive objective is
\begin{equation}
\begin{split}
\mathcal{L}_{\mathrm{pheno}}^{(k)}
=
&-\frac{1}{|\mathcal{B}|}
\sum_{i\in\mathcal{B}}
\frac{1}{|\mathcal{P}_k(i)|}
\sum_{j\in\mathcal{P}_k(i)}
\\
&\log
\frac{
\exp\left(
\mathrm{sim}\left(
\mathbf{r}_i^{(k)},
\mathbf{r}_j^{(k)}
\right)/\tau
\right)
}{
\displaystyle
\sum_{a\in\mathcal{B}\setminus\{i\}}
\exp\left(
\mathrm{sim}\left(
\mathbf{r}_i^{(k)},
\mathbf{r}_a^{(k)}
\right)/\tau
\right)
}.
\end{split}
\end{equation}
Here, $\mathrm{sim}(\cdot,\cdot)$ denotes cosine similarity and
$\tau$ is the temperature parameter. The phenotype-aware sampler
ensures that $|\mathcal{P}_k(i)|\geq 1$ for every anchor. Following supervised contrastive learning~\cite{khosla2020supervised}, the denominator
includes all non-anchor admissions in the mini-batch, including
positive examples.

\subsubsection{Joint Training Objective}

The final task-specific objective combines admission-level risk
prediction with phenotype-guided representation learning:
\[
\mathcal{L}^{(k)}
=
\mathcal{L}_{\mathrm{pred}}^{(k)}
+
\lambda\mathcal{L}_{\mathrm{pheno}}^{(k)},
\]
where $\lambda$ controls the contribution of the phenotype-guided
contrastive objective. The temperature $\tau$ and loss weight
$\lambda$ are selected independently for each dataset--task setting,
as described in Section~\ref{sec:implementation_details}.

%% file: 3.exp.tex
\subsection{Datasets}
\label{sec:datasets}
We evaluated \modelname{} on three de-identified public clinical datasets:
MIMIC-IV~\cite{PhysioNet-mimiciv-3.1}, the eICU Collaborative Research Database~\cite{eicu}, and MIMIC-III~\cite{mimic-iii}. 
Admissions
shorter than two days were excluded, and each dataset was split at the patient
level into training, validation, and test sets using a 75:10:15 ratio. The unit
of analysis was a hospital admission for MIMIC-IV and MIMIC-III and an ICU stay
for eICU.

We considered three binary tasks. In-hospital mortality was defined as whether
the patient died during the index hospitalization and was predicted using data
recorded through the final hospital day. Long LOS was defined as a total
hospital stay of at least seven days and was predicted using data from the first
48 hours after admission. Thirty-day readmission was defined as whether a
subsequent admission occurred within 30 days after discharge and was predicted
using data recorded through the final hospital day.

Demographics and diagnosis-derived information were encoded as global context,
while time-stamped clinical events were organized as local events.
The available local modalities differed across datasets. For Long LOS, local
inputs were restricted to the first two hospital days, and later events and
terminal discharge summaries were excluded.

\indent \textbf{MIMIC-IV and MIMIC-III.}
The two MIMIC datasets contain structured EHR events, radiology reports, and discharge summaries. Following prior multimodal clinical prediction setups, we used \texttt{admissions}~\cite{mmstgnn,crl-mmnar,muse,adacomed,multimodalpred}, \texttt{diagnoses\_icd}~\cite{mmstgnn, crl-mmnar, muse}, \texttt{labevents}~\cite{mmstgnn, crl-mmnar, muse, adacomed, multimodalpred}, \texttt{prescriptions}~\cite{mmstgnn, muse} as validated in prior structured EHR setups, alongside radiology reports~\cite{adacomed, crl-mmnar, multimodalpred}, and discharge summaries~\cite{crl-mmnar, muse} to capture the corresponding clinical text modalities. MIMIC-IV was additionally linked to MIMIC-CXR~\cite{PhysioNet-mimic-cxr-2.1.0} using patient- and study-level identifiers to incorporate chest x-rays. The final cohorts contained 10,387/1,400/2,032 training/validation/test admissions for MIMIC-IV and 34,342/4,525/6,800 for MIMIC-III.

\indent \textbf{eICU.}
eICU provided an ICU-specific setting based on structured EHR events. To maintain consistency with the setup used by
MUSE+~\cite{muse}, we constructed the eICU cohort
using the \texttt{patient}, \texttt{diagnosis}, \texttt{treatment}, \texttt{medication}, \texttt{lab}, and \texttt{apacheApsVar} tables. APACHE-related severity summaries from \texttt{apacheApsVar} were included as global context, while medications, laboratory measurements, and treatments from the remaining tables were used as local inputs. The final cohort contained 89,805/11,974/17,962 training/validation/test ICU stays.

\subsection{Implementation Details}
\label{sec:implementation_details}

\modelname{} first encodes each available modality using a pretrained
modality-specific encoder, projects the resulting representations into a shared
model space, and applies the masked Transformer encoder with the TGL attention mask described in Section~\ref{sec:TGL}. Chest x-rays were resized to
$224\times224$ pixels and encoded using a medical vision encoder~\cite{eva-x},
producing 768-dimensional image representations. Pretrained vision and language encoders were kept frozen throughout training; only the modality-specific projection layers, the temporal global–local trajectory encoder, and the task-specific prediction and contrastive heads were updated.

Structured EHR inputs were converted into value-preserving natural-language descriptors for numeric measurements (e.g., \texttt{Hematocrit: 32.1 g/dL}), preserving their originally recorded precision and units without additional rounding, normalization, or outlier clipping. Categorical concepts were represented using code-based descriptors (e.g., \texttt{I50.9} $\rightarrow$ \texttt{Heart failure}). The resulting descriptors were encoded using a clinical language encoder~\cite{modernbert}. Radiology reports and discharge summaries, when available for the corresponding
task, were encoded using the same language encoder with a maximum sequence length of 512 tokens. The 768-dimensional \texttt{[CLS]} representation was
used for each textual input. Modality-specific projection layers mapped all
encoded inputs into the shared hidden dimension of \modelname{}.

Models were trained for up to 30 epochs using AdamW with a warmup--cosine
learning-rate schedule. Early stopping was performed according to validation
performance. The contrastive temperature $\tau$ and loss weight $\lambda$ were
selected independently for each dataset--task setting using the validation set.
Across all settings, the selected values averaged $\lambda=0.42$ and
$\tau=0.09$.

We report accuracy (ACC), area under the receiver operating characteristic
curve (AUROC), area under the precision--recall curve (AUPRC), and macro-F1
(mF1). ACC and mF1 were computed from sigmoid-transformed positive-class
probabilities using the decision threshold selected on the validation set.
95\%  confidence intervals were estimated using 2,000 bootstrap
resamples of the test set.

\input{table/table_fix_all}

\subsection{Comparison with State-of-the-Art Methods}
\label{sec:sota_compare}
Table~\ref{tab:all_datasets_split} compares \modelname{} with multimodal
clinical prediction baselines grouped by their primary modeling emphasis:
fusion- and relation-centered methods
(Large-Ensemble~\cite{adacomed}, MultimodalPred~\cite{multimodalpred}, MM-STGNN~\cite{mmstgnn}, and AdaCoMed~\cite{adacomed}) and
missing-modality-aware methods
(M3Care~\cite{m3care}, MUSE+~\cite{muse}, and CRL-MMNAR~\cite{crl-mmnar}). Across three datasets and three tasks,
\modelname{} achieved the highest AUROC in all nine dataset--task settings and
the highest ACC in eight, demonstrating consistent discrimination across
heterogeneous inpatient and ICU cohorts.

For in-hospital mortality, \modelname{} achieved AUROCs of
0.9804, 0.9490, and 0.9770 on MIMIC-IV, eICU, and MIMIC-III, respectively.
These values exceeded the second-best results by 0.0015, 0.0009, and 0.0018.
Although the AUROC gains were consistent, CRL-MMNAR achieved higher AUPRC and
mF1 in several mortality settings, indicating that the relative performance
varied across discrimination, precision--recall, and threshold-dependent metrics.

For Long LOS prediction from the first 48 hours, \modelname{} achieved the
highest AUROC on MIMIC-IV, eICU, and MIMIC-III
(0.7675, 0.7971, and 0.7864), improving over the second-best method by
0.0125, 0.0360, and 0.0113, respectively. \modelname{} also achieved the
highest mF1 on MIMIC-IV and the highest AUPRC on eICU and MIMIC-III. These
results indicate that the proposed admission representation captures useful
early admission patterns associated with prolonged hospitalization.

For 30-day readmission, \modelname{} achieved the highest AUROC on MIMIC-IV,
eICU, and MIMIC-III
(0.8371, 0.7584, and 0.8306), exceeding the second-best method by
0.0245, 0.0694, and 0.0064, respectively. It also achieved the highest AUPRC
on MIMIC-III. Although AUPRC and mF1 varied across cohorts, the AUROC results
showed consistent gains for discharge-time readmission prediction.

Overall, \modelname{} produced the most consistent gains in ranking discrimination across datasets and tasks, while improvements in AUPRC and mF1 were less uniform.

\subsection{Ablation Studies}
The following ablations isolate \modelname{} design choices under fixed architectures and encoders, complementing the baseline comparison in Section~\ref{sec:sota_compare}.

\subsubsection{Contribution of Input Modalities}
\input{table/modality_exp}

We conducted modality ablations on MIMIC-IV and MIMIC-III to assess the
contribution of individual input modalities and their combinations
(Table~\ref{tab:merged_ablation_study}). MIMIC-IV includes chest x-rays,
structured EHR events, radiology reports, and discharge summaries, whereas
MIMIC-III includes structured EHR events, radiology reports, and discharge
summaries. Discharge-summary configurations were excluded from Long LOS because
this task uses only data recorded during the first 48 hours.

Structured EHR events provided the strongest unimodal AUROC for mortality and
readmission in both datasets. For Long LOS, the strongest unimodal modality
differed by dataset, with chest x-rays performing best on MIMIC-IV and
radiology reports performing best on MIMIC-III. Multimodal configurations
generally improved AUROC over unimodal settings, especially when structured EHR
events were combined with imaging or radiology reports. Discharge summaries
provided useful terminal context for mortality and readmission, but showed
limited standalone performance and were not applicable to the first-48-hour
Long LOS setting.

The task-specific complete \modelname{} configuration achieved the highest
AUROC for all tasks on both datasets: 0.9804/0.7675/0.8371 on MIMIC-IV and
0.9770/0.7864/0.8306 on MIMIC-III for mortality, Long LOS, and readmission,
respectively. Although some partial configurations achieved higher AUPRC or
macro-F1 in specific settings, the consistent AUROC gains support the value of
organizing available modalities within the proposed trajectory representation.

\subsubsection{TGL Attention-Mask Ablation}
\input{table/mask_ablation}

We evaluated the TGL attention mask against two alternative attention masks: All-Causal and Full-Attention. 
The All-Causal mask imposes a single causal order over all tokens, including global context, whereas the Full-Attention mask permits bidirectional interactions across all hospital days. These alternatives represent strictly sequential and temporally unconstrained encoding, respectively.

As shown in Table~\ref{tab:context_ablation}, the TGL mask achieved
the highest AUROC across mortality, Long LOS, and readmission
(0.9804, 0.7675, and 0.8371), compared with
0.9778/0.7583/0.8271 for All-Causal and
0.9576/0.7462/0.8042 for Full-Attention. 
It also achieved the highest AUPRC for mortality and Long LOS. For readmission, its AUPRC matched All-Causal (0.6804) and exceeded Full-Attention (0.6323). The TGL mask also achieved the highest mF1 for mortality and readmission, while All-Causal yielded a higher mF1 for Long LOS (0.6967 vs.\ 0.6750). Overall, TGL provided consistent gains in AUROC and AUPRC, although its advantage was not uniform across threshold-dependent metrics.

\input{table/contrast_bce}
\subsubsection{Phenotype-Guided Contrastive Objective}

We evaluated the contribution of the phenotype-guided contrastive objective by comparing the full joint objective $\mathcal{L}^{(k)}$ with the prediction-only objective $\mathcal{L}_{\mathrm{pred}}^{(k)}$, while keeping the trajectory encoder, prediction head, and input configuration unchanged. Table~\ref{tab:pcl_bce} reports the $L_{pred}$-only performance and the change obtained by adding $L_{pheno}$, defined as \[
\Delta_{\mathrm{pheno}}
=
\operatorname{Metric}\!\left(\mathcal{L}^{(k)}\right)
-
\operatorname{Metric}\!\left(
\mathcal{L}_{\mathrm{pred}}^{(k)}
\right)
\]

$L_{pheno}$ was trained using all diagnosis-derived phenotype groups in each dataset. For phenotype-stratified evaluation, each test admission was assigned to a single phenotype to form mutually exclusive groups. The six most frequent phenotypes were selected separately for each dataset, and the reported average is the unweighted mean across these six groups.

Adding $L_{pheno}$ improved mean phenotype-stratified AUROC in all nine dataset--task
settings. The largest gains occurred for eICU mortality ($+0.0274$) and Long LOS ($+0.0164$). Mean AUPRC improved in seven of the
nine settings, including mortality on eICU
($+0.0640$) and MIMIC-IV ($+0.0467$). The exceptions were MIMIC-IV readmission and eICU Long LOS, for
which mean AUPRC decreased slightly.

The effects varied across phenotypes. For example, eICU Trauma and Poisoning
showed an increase of $+0.1476$ in mortality AUPRC, whereas some smaller
phenotype groups showed limited or negative changes. Thus, the joint objective consistently improved overall phenotype-stratified AUROC, although the gains were not uniform across all metrics or phenotype groups.

\begin{figure*}[t]
    \centering
    \scriptsize
    \setlength{\tabcolsep}{1pt}
    \renewcommand{\arraystretch}{0.92}

    \begin{tabular}{@{}c c c c@{}}
        &
        \textbf{Mortality} &
        \textbf{Long LOS} &
        \textbf{Readmission}
        \\[-12pt]

        \rotatebox{90}{
            \parbox{0.2\textwidth}{\centering
            \textbf{Joint objective}
            }
        }
        &
        \includegraphics[width=0.298\textwidth]{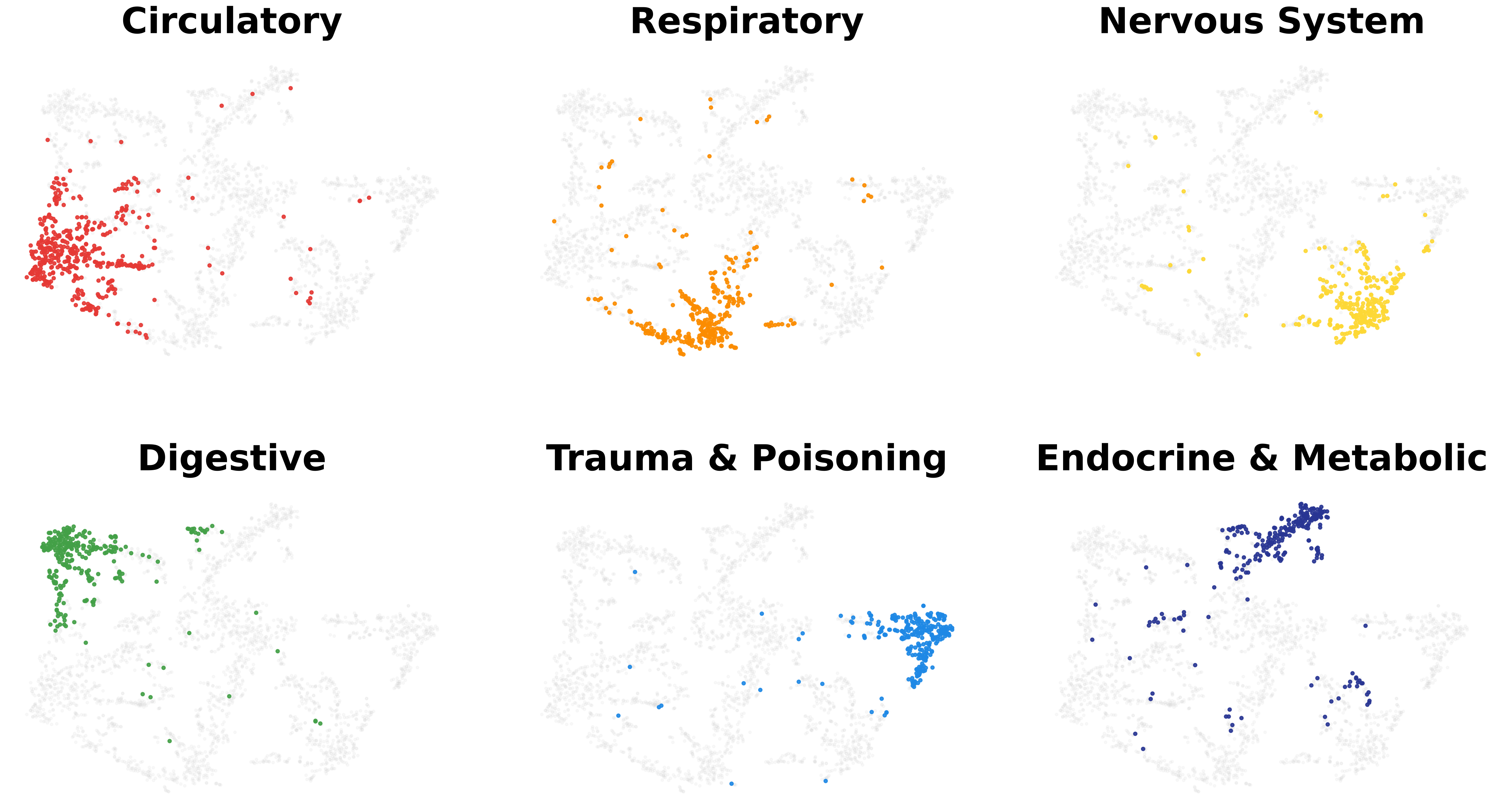}
        &
        \includegraphics[width=0.298\textwidth]{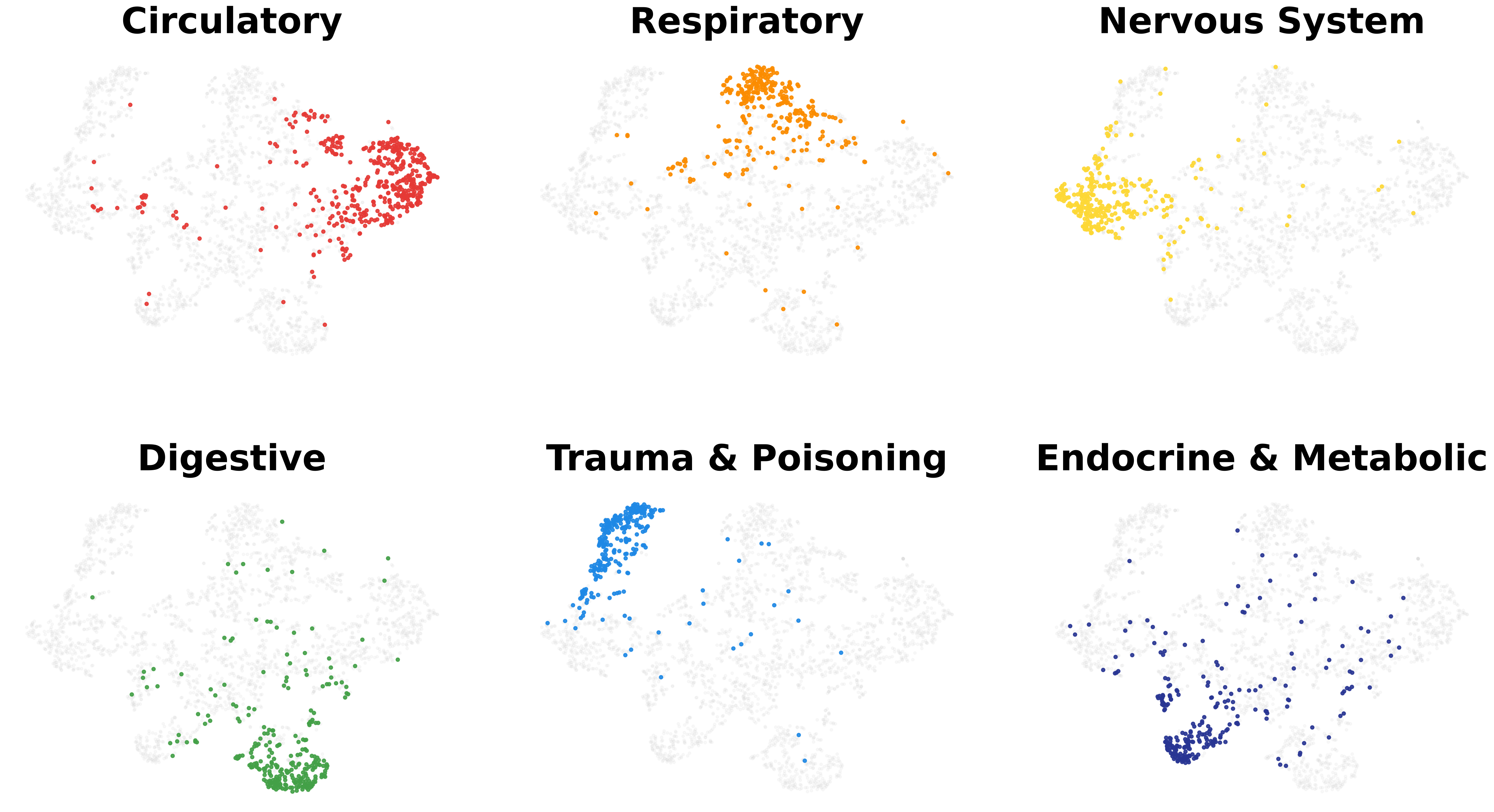}
        &
        \includegraphics[width=0.298\textwidth]{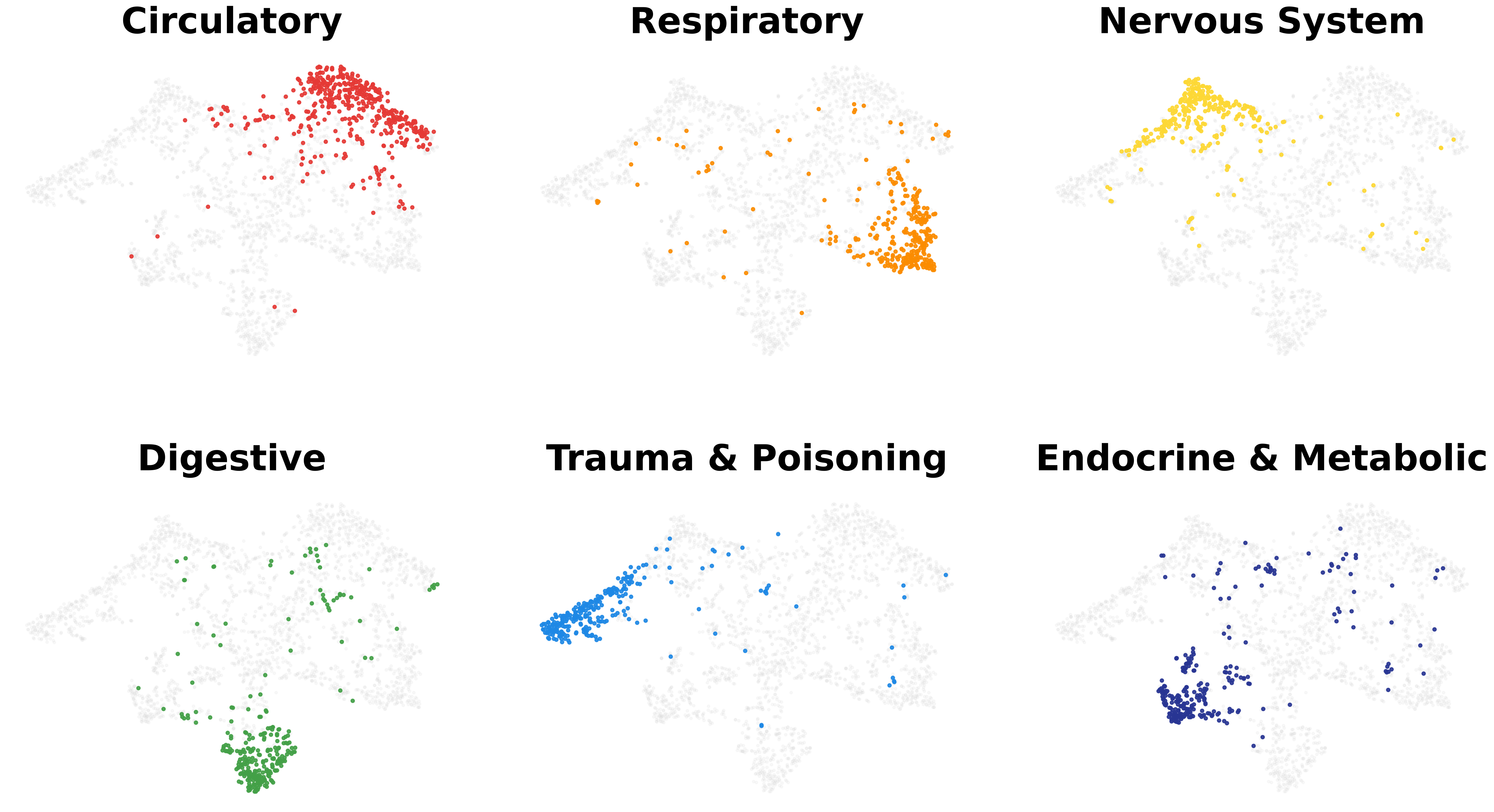}
        \\[-5pt]

        \rotatebox{90}{
            \parbox{0.2\textwidth}{
                \centering
                \textbf{Prediction-only}
            }
        }
        &
        \includegraphics[width=0.298\textwidth]{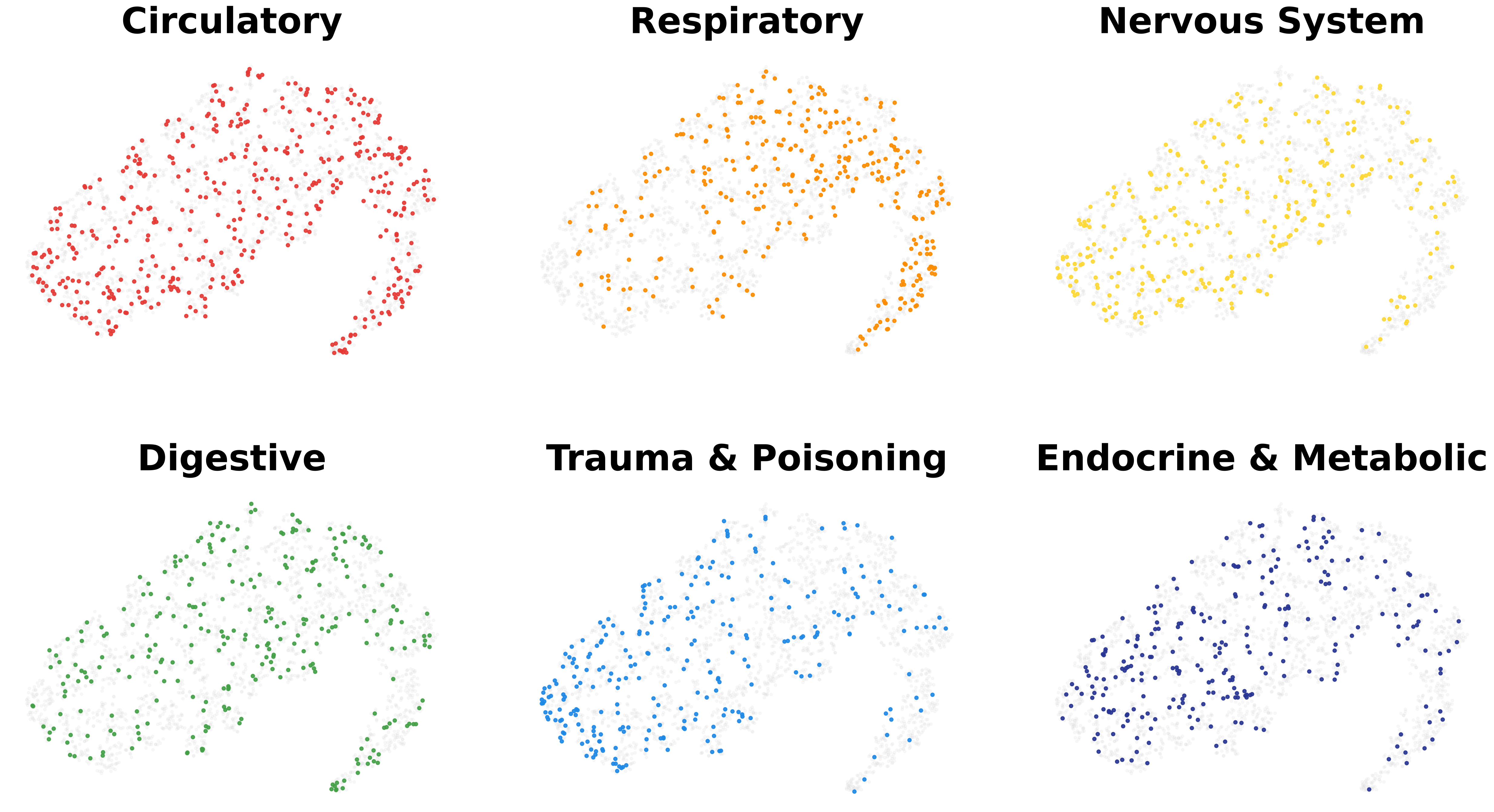}
        &
        \includegraphics[width=0.298\textwidth]{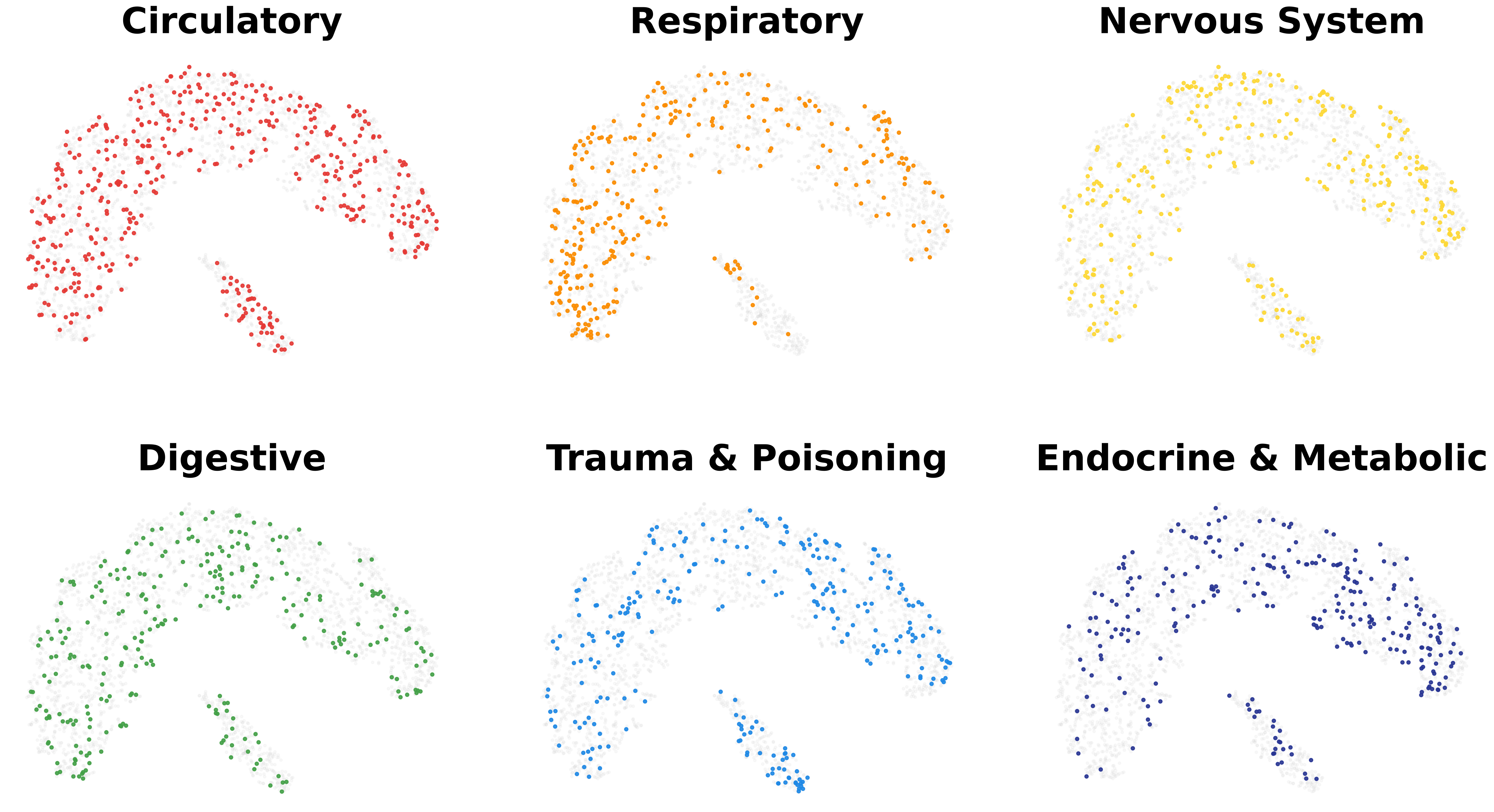}
        &
        \includegraphics[width=0.298\textwidth]{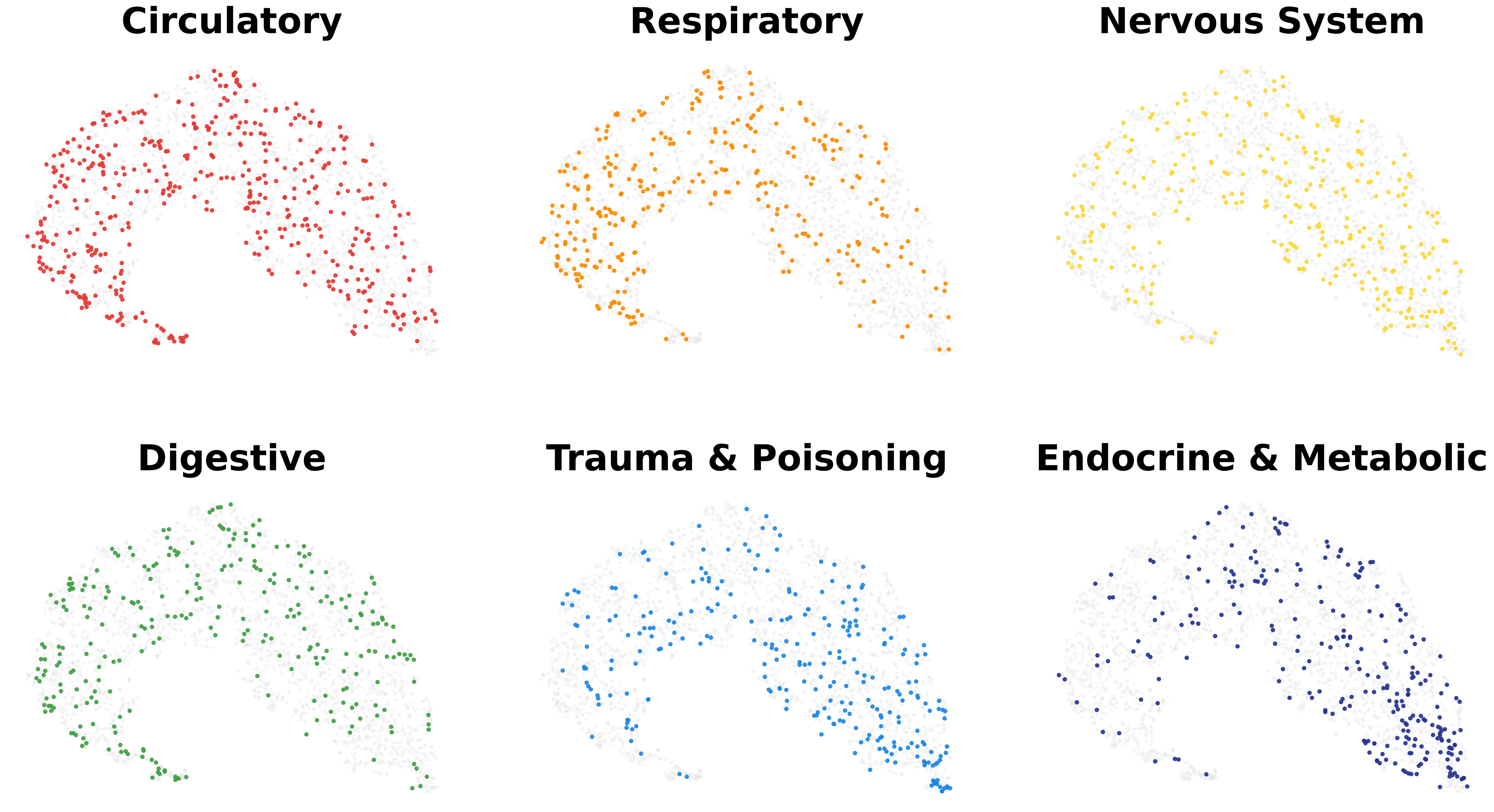}
    \end{tabular}

    \caption{
        t-SNE~\cite{van2008visualizing} visualization of task-specific admission representations
        learned with the joint objective
        $\mathcal{L}_{\mathrm{pred}}^{(k)}
        +\lambda\mathcal{L}_{\mathrm{pheno}}^{(k)}$
        and the prediction-only objective
        $\mathcal{L}_{\mathrm{pred}}^{(k)}$.
        Points are colored by diagnosis-derived phenotype group.
    }
    \label{fig:tsne_comparison_matrix}
\end{figure*}

Figure~\ref{fig:tsne_comparison_matrix} provides a complementary qualitative view of the learned representations. Compared with $L_{pred}$ alone, adding $L_{pheno}$ produces greater phenotype-level organization across mortality, Long LOS, and readmission, consistent with its role as a phenotype-aware representation regularizer.

\begin{figure*}[p]
    \centering
    \captionsetup{
        font=small,
        skip=3pt
    }
    \captionsetup[subfigure]{
        font=scriptsize,
        labelformat=parens,
        labelsep=space,
        justification=centering,
        singlelinecheck=true,
        skip=1pt,
        margin=0pt
    }

    \begin{subfigure}[t]{\textwidth}
        \centering
        \includegraphics[
            width=\textwidth,
            height=0.45\textheight,
            keepaspectratio
        ]{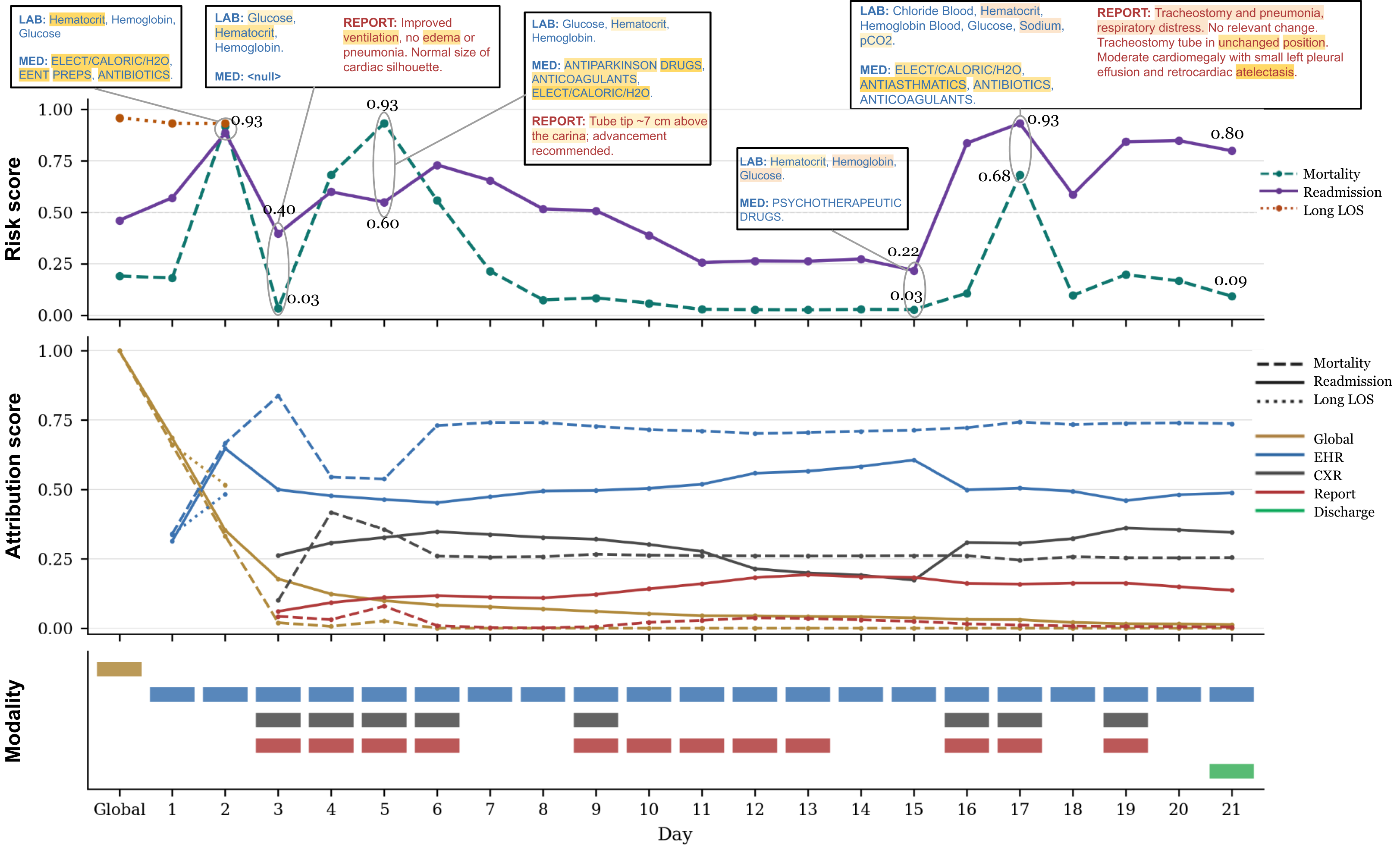}
        \caption{Respiratory phenotype with septicemia.}
        \label{fig:case_respiratory}
    \end{subfigure}


    \begin{subfigure}[t]{\textwidth}
        \centering
        \includegraphics[
            width=\textwidth,
            height=0.45\textheight,
            keepaspectratio
        ]{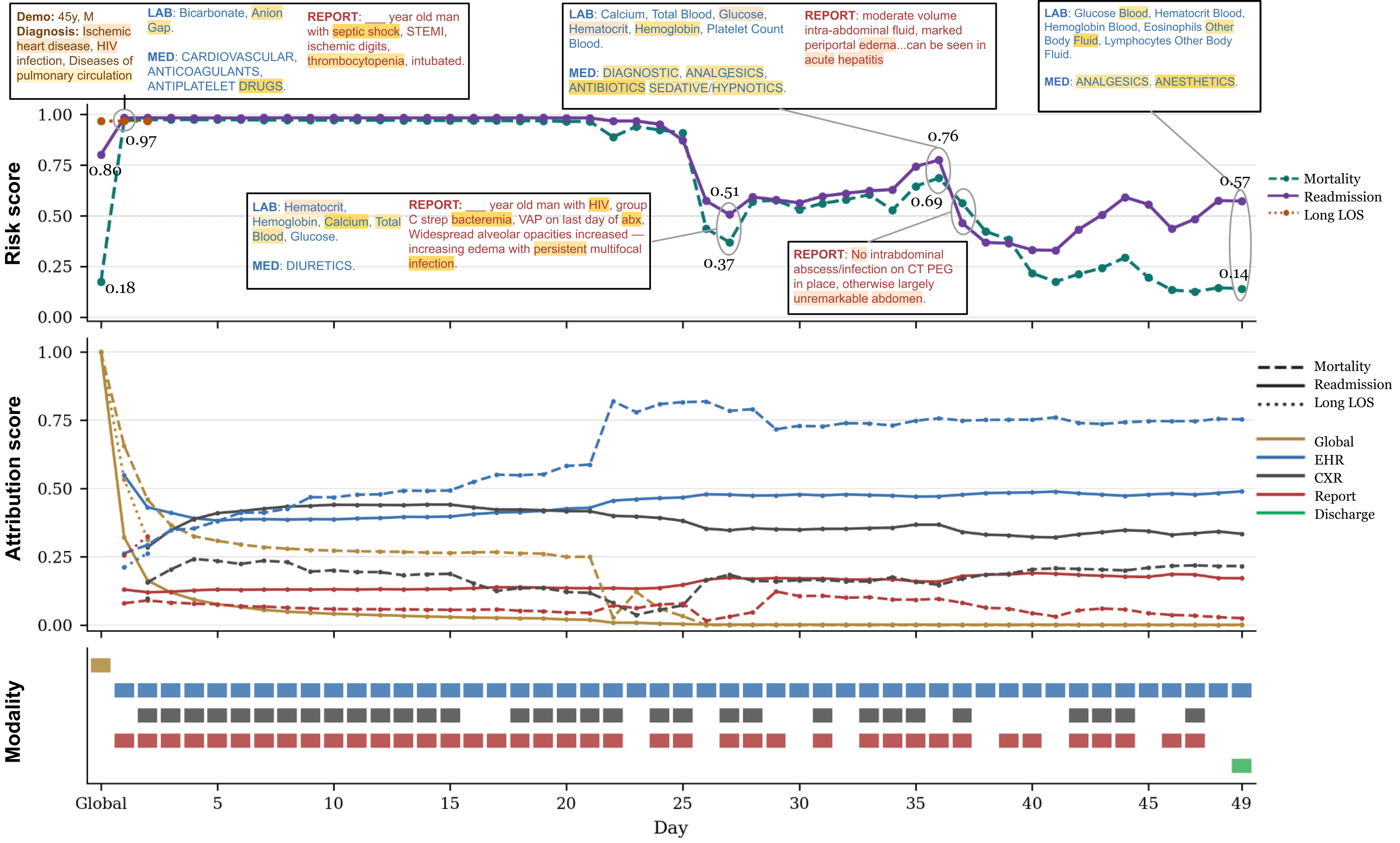}
        \caption{Circulatory phenotype with heart failure.}
        \label{fig:case_circulatory}
    \end{subfigure}


    \caption{
        Trajectory-level interpretability for representative patients from two diagnosis-derived phenotypes. Prefix-level risk estimates are obtained through repeated inference using global context and local clinical events observed through each hospital day. Gradient$\times$input attribution scores are aggregated by hospital day and modality for structured EHR, chest x-rays, radiology reports, and discharge summaries.}
    \label{fig:trajectory_cases}
\end{figure*}

\subsection{Trajectory-Level Interpretability}
\label{sec:trajectory}
Figure~\ref{fig:trajectory_cases} shows how \modelname{} links admission-level mortality, readmission, and Long LOS predictions to day- and modality-specific clinical events within a single patient's stay. Day-level trajectories are obtained via repeated prefix inference: at each hospital day $t$, the model receives all clinical events recorded up to $t$ and produces task-specific estimates $\hat{p}_{i,t}^{(k)}$. Attribution was computed as the elementwise product of each input embedding and the gradient of the task probability with respect to that embedding. The scores were summed over embedding dimensions and aggregated by hospital day and modality for EHR, chest x-rays, radiology reports, and discharge summaries. This analysis is a qualitative interpretation of model behavior, not a quantitative test of clinical causality.

In the respiratory septicemia case (Figure~\ref{fig:case_respiratory}, 21-day admission), a patient with oropharyngeal cancer and a tracheostomy underwent airway evaluation (day 4), tube-position monitoring (days 5--6), a pleural effusion (day 16), and pseudomonas pneumonia (day 19) before stabilizing. The Long LOS prediction is high from admission (0.96) and stays high through day 2 (0.93), reflecting reliance on admission-level diagnosis codes. Mortality and readmission instead track new findings: both rise on day 2 with new labs and medications (attribution shifting to EHR), then fall on day 3 as a report notes \textit{``improved ventilation, no edema or pneumonia''} (attribution shifting to the report). Both rise again on days 5--6 with a new anticoagulant and a report flagging tube malposition, then stay low through days 9--15. A day-16 pleural effusion raises readmission sharply (0.22$\to$0.84) but mortality only slightly (0.03$\to$0.11). Both rise again on day 17 despite a negative report, coinciding instead with new electrolyte/pCO2 abnormalities and added medications. Attribution consistently follows whichever modality carries new information at each step.

In the circulatory case (Figure~\ref{fig:case_circulatory}, 49-day admission), a patient with HIV and septic shock complicated by rhabdomyolysis, tracheostomy, and PEG placement developed ventilator-associated pneumonia, with worsening consolidation (day 24), improvement after diuresis (days 25--26), recurrent fever/bacteremia (day 27), a hepatitis-like finding (day 36), and a deep-vein thrombosis (day 40). As in the respiratory case, Long LOS is high from admission (0.97) and stays high through day 2 (0.97). Mortality and readmission remain high (0.97--0.98) through day 21, then both decrease sharply by day 27 (0.91$\to$0.37 and 0.87$\to$0.51), following a report of \textit{``rapidly improved''} opacities on day 25 and new diuretics on day 26. Risk peaks again on day 36 with periportal edema/hepatitis findings and new antibiotics/sedatives, then falls once a repeat CT is unremarkable. On day 40, a new DVT and anticoagulants do not raise risk, consistent with the report modality carrying only 3--4\% attribution versus over 75\% for EHR.

Together, these cases show that trajectory-level attribution connects admission-level predictions to day- and modality-specific clinical events, revealing distinct pathways and task-specific sensitivities underlying mortality, Long LOS, and readmission risk.

%% file: table/table_fix_all.tex
\begin{sidewaystable*}
\centering
\caption{Performance comparison of the proposed model and baseline methods on three inpatient and critical-care datasets (MIMIC-IV, eICU, and MIMIC-III) for mortality, Long LOS, and readmission prediction tasks. Values in brackets indicate 95\% confidence intervals, calculated via bootstrapping with 2,000 resamples.}
\label{tab:all_datasets_split}

\vspace{0.2cm}
\textbf{(a) MIMIC-IV}

\vspace{0.1cm}
\setlength{\tabcolsep}{3pt} 
\renewcommand{\arraystretch}{1.4}
\resizebox{\linewidth}{!}{
\begin{tabular}{l *{4}{c} *{4}{c} *{4}{c}}
  \toprule
  \multirow{2}{*}{\textbf{Method}} & \multicolumn{4}{c}{\textbf{Mortality}} & \multicolumn{4}{c}{\textbf{Long LOS}} & \multicolumn{4}{c}{\textbf{Readmission}} \\
  \cmidrule(lr){2-5} \cmidrule(lr){6-9} \cmidrule(lr){10-13}
  & \textbf{ACC} & \textbf{AUROC} & \textbf{AUPRC} & \textbf{mF1} 
  & \textbf{ACC} & \textbf{AUROC} & \textbf{AUPRC} & \textbf{mF1} 
  & \textbf{ACC} & \textbf{AUROC} & \textbf{AUPRC} & \textbf{mF1} \\
  \midrule
  
  Large-Ensemble 
  & 0.8521 {\scriptsize [0.8298, 0.8744]} & 0.6564 {\scriptsize [0.6074, 0.7060]} & 0.2567 {\scriptsize [0.1978, 0.3257]} & 0.4919 {\scriptsize [0.4669, 0.5233]} & 0.7345 {\scriptsize [0.7045, 0.7603]} & 0.7205 {\scriptsize [0.6852, 0.7542]} & 0.7677 {\scriptsize [0.7406, 0.7906]} & 0.5528 {\scriptsize [0.5148, 0.5874]} & 0.8136 {\scriptsize [0.7893, 0.8379]} & 0.5626 {\scriptsize [0.5133, 0.6082]} & 0.2297 {\scriptsize [0.1862, 0.2797]} & 0.4486 {\scriptsize [0.4411, 0.4559]} \\
  
  MultimodalPred 
  & 0.3931 {\scriptsize [0.3627, 0.4235]} & 0.6213 {\scriptsize [0.5752, 0.6694]} & 0.2069 {\scriptsize [0.1620, 0.2643]} & 0.3792 {\scriptsize [0.3497, 0.4091]} & 0.6147 {\scriptsize [0.5837, 0.6457]} & 0.5447 {\scriptsize [0.5032, 0.5862]} & 0.7532 {\scriptsize [0.7156, 0.7878]} & 0.5309 {\scriptsize [0.4996, 0.5640]} & 0.8136 {\scriptsize [0.7893, 0.8379]} & 0.6308 {\scriptsize [0.5869, 0.6746]} & 0.2907 {\scriptsize [0.2365, 0.3533]} & 0.4970 {\scriptsize [0.4690, 0.5313]} \\
  
  MM-STGNN 
  & 0.9321 {\scriptsize [0.9213, 0.9429]} & 0.9515 {\scriptsize [0.9367, 0.9641]} & 0.7878 {\scriptsize [0.7363, 0.8341]} & 0.8370 {\scriptsize [0.8101, 0.8618]} & 0.6430 {\scriptsize [0.6229, 0.6623]} & \underline{0.7550 {\scriptsize [0.7349, 0.7745]}} & 0.7331 {\scriptsize [0.7037, 0.7620]} & 0.6144 {\scriptsize [0.5928, 0.6353]} & 0.8553 {\scriptsize [0.8396, 0.8711]} & 0.7938 {\scriptsize [0.7636, 0.8237]} & 0.6262 {\scriptsize [0.5772, 0.6717]} & 0.7385 {\scriptsize [0.7111, 0.7637]} \\
  
  M3care 
  & 0.8917 {\scriptsize [0.8784, 0.9050]} & 0.9733 {\scriptsize [0.9617, 0.9826]} & 0.8567 {\scriptsize [0.8056, 0.8988]} & 0.7990 {\scriptsize [0.7748, 0.8220]} & 0.5917 {\scriptsize [0.5606, 0.6217]} & 0.6897 {\scriptsize [0.6546, 0.7225]} & 0.7588 {\scriptsize [0.7212, 0.7940]} & 0.5909 {\scriptsize [0.5597, 0.6210]} & 0.8740 {\scriptsize [0.8583, 0.8878]} & 0.7940 {\scriptsize [0.7644, 0.8234]} & 0.6338 {\scriptsize [0.5866, 0.6793]} & 0.7357 {\scriptsize [0.7055, 0.7621]} \\
  
  MUSE+ 
  & 0.9350 {\scriptsize [0.9242, 0.9454]} & 0.9537 {\scriptsize [0.9364, 0.9686]} & 0.8023 {\scriptsize [0.7450, 0.8528]} & 0.8525 {\scriptsize [0.8276, 0.8756]} & \textbf{0.7569 {\scriptsize [0.7382, 0.7756]}} & 0.7421 {\scriptsize [0.7160, 0.7658]} & \underline{0.8571 {\scriptsize [0.8358, 0.8783]}} & \underline{0.6669 {\scriptsize [0.6435, 0.6910]}} & 0.8652 {\scriptsize [0.8499, 0.8799]} & 0.7913 {\scriptsize [0.7604, 0.8201]} & 0.6441 {\scriptsize [0.5945, 0.6892]} & \underline{0.7592 {\scriptsize [0.7329, 0.7820]}} \\
  
  AdaCoMed 
  & 0.5765 {\scriptsize [0.5451, 0.6069]} & 0.6730 {\scriptsize [0.6266, 0.7185]} & 0.2499 {\scriptsize [0.1952, 0.3173]} & 0.4986 {\scriptsize [0.4670, 0.5295]} & 0.7324 {\scriptsize [0.7045, 0.7583]} & 0.7166 {\scriptsize [0.6809, 0.7500]} & \textbf{0.8604 {\scriptsize [0.8311, 0.8867]}} & 0.4933 {\scriptsize [0.4631, 0.5247]} & \underline{0.8781 {\scriptsize [0.8598, 0.9065]}} & 0.6522 {\scriptsize [0.6105, 0.6951]} & 0.2975 {\scriptsize [0.2453, 0.3570]} & 0.3173 {\scriptsize [0.2886, 0.3454]} \\
  
  CRL-MMNAR 
  & \underline{0.9485 {\scriptsize [0.9406, 0.9559]}} & \underline{0.9789 {\scriptsize [0.9682, 0.9878]}} & \textbf{0.9061 {\scriptsize [0.8618, 0.9401]}} & \textbf{0.9161 {\scriptsize [0.8951, 0.9359]}} & 0.6863 {\scriptsize [0.6725, 0.6996]} & 0.7294 {\scriptsize [0.7171, 0.7414]} & 0.7680 {\scriptsize [0.7606, 0.7746]} & 0.6577 {\scriptsize [0.6410, 0.6742]} & 0.8672 {\scriptsize [0.8254, 0.8980]} & \underline{0.8126 {\scriptsize [0.7920, 0.8320]}} & \textbf{0.7367 {\scriptsize [0.7013, 0.7693]}} & \textbf{0.7681 {\scriptsize [0.7474, 0.7879]}} \\
  
  \midrule 
  
  \modelname{} 
  & \textbf{0.9528 {\scriptsize [0.9424, 0.9626]}} & \textbf{0.9804 {\scriptsize [0.9684, 0.9893]}} & \underline{0.8605 {\scriptsize [0.8117, 0.9019]}} & \underline{0.8886 {\scriptsize [0.8657, 0.9095]}} & \underline{0.7409 {\scriptsize [0.7179, 0.7640]}} & \textbf{0.7675 {\scriptsize [0.7418, 0.7929]}} & 0.8428 {\scriptsize [0.8152, 0.8684]} & \textbf{0.6750 {\scriptsize [0.6475, 0.7019]}} & \textbf{0.8804 {\scriptsize [0.8656, 0.8947]}} & \textbf{0.8371 {\scriptsize [0.8096, 0.8636]}} & \underline{0.6804 {\scriptsize [0.6342, 0.7248]}} & 0.7567 {\scriptsize [0.7315, 0.7806]} \\
  \bottomrule
\end{tabular}
}

\vspace{0.4cm}
\textbf{(b) eICU}

\vspace{0.1cm}
\resizebox{\linewidth}{!}{
\begin{tabular}{l *{4}{c} *{4}{c} *{4}{c}}
  \toprule
  \multirow{2}{*}{\textbf{Method}} & \multicolumn{4}{c}{\textbf{Mortality}} & \multicolumn{4}{c}{\textbf{Long LOS}} & \multicolumn{4}{c}{\textbf{Readmission}} \\
  \cmidrule(lr){2-5} \cmidrule(lr){6-9} \cmidrule(lr){10-13}
  & \textbf{ACC} & \textbf{AUROC} & \textbf{AUPRC} & \textbf{mF1} 
  & \textbf{ACC} & \textbf{AUROC} & \textbf{AUPRC} & \textbf{mF1} 
  & \textbf{ACC} & \textbf{AUROC} & \textbf{AUPRC} & \textbf{mF1} \\
  \midrule
  
  Large-Ensemble 
  & 0.8614 {\scriptsize [0.8561, 0.8663]} & 0.9249 {\scriptsize [0.9155, 0.9335]} & 0.5261 {\scriptsize [0.4877, 0.5605]} & 0.6363 {\scriptsize [0.6254, 0.6464]} & \underline{0.9441 {\scriptsize [0.9410, 0.9474]}} & 0.7502 {\scriptsize [0.7353, 0.7637]} & 0.1427 {\scriptsize [0.1292, 0.1585]} & 0.4866 {\scriptsize [0.4850, 0.4891]} & 0.7802 {\scriptsize [0.7736, 0.7871]} & 0.6795 {\scriptsize [0.6661, 0.6925]} & 0.1848 {\scriptsize [0.1717, 0.1997]} & 0.2730 {\scriptsize [0.2668, 0.2797]} \\
  
  MultimodalPred 
  & 0.9452 {\scriptsize [0.9416, 0.9484]} & 0.9102 {\scriptsize [0.8998, 0.9198]} & 0.4553 {\scriptsize [0.4160, 0.4918]} & 0.7163 {\scriptsize [0.7005, 0.7304]} & 0.9240 {\scriptsize [0.9213, 0.9278]} & 0.7156 {\scriptsize [0.6986, 0.7330]} & 0.1341 {\scriptsize [0.1214, 0.1488]} & 0.4757 {\scriptsize [0.4749, 0.4766]} & 0.8088 {\scriptsize [0.8044, 0.8134]} & 0.6540 {\scriptsize [0.6411, 0.6666]} & 0.1631 {\scriptsize [0.1517, 0.1755]} & 0.5021 {\scriptsize [0.4982, 0.5062]} \\
  
  MM-STGNN 
  & 0.9124 {\scriptsize [0.9009, 0.9236]} & 0.7880 {\scriptsize [0.7432, 0.8317]} & 0.1472 {\scriptsize [0.0915, 0.2198]} & 0.5715 {\scriptsize [0.5390, 0.6050]} & 0.8560 {\scriptsize [0.8413, 0.8711]} & 0.6851 {\scriptsize [0.6388, 0.7277]} & 0.1072 {\scriptsize [0.0796, 0.1449]} & \underline{0.5366 {\scriptsize [0.5114, 0.5632]}} & 0.8089 {\scriptsize [0.7929, 0.8249]} & 0.5942 {\scriptsize [0.5546, 0.6342]} & 0.1375 {\scriptsize [0.1132, 0.1687]} & 0.5356 {\scriptsize [0.5119, 0.5598]} \\
  
  M3care 
  & 0.9326 {\scriptsize [0.9289, 0.9363]} & 0.9477 {\scriptsize [0.9401, 0.9549]} & \underline{0.6167 {\scriptsize [0.5806, 0.6538]}} & 0.7381 {\scriptsize [0.7254, 0.7502]} & 0.7677 {\scriptsize [0.7615, 0.7742]} & \underline{0.7611 {\scriptsize [0.7472, 0.7750]}} & 0.1452 {\scriptsize [0.1310, 0.1606]} & \textbf{0.5415 {\scriptsize [0.5331, 0.5496]}} & 0.7862 {\scriptsize [0.7801, 0.7922]} & 0.6353 {\scriptsize [0.6233, 0.6464]} & 0.2214 {\scriptsize [0.2094, 0.2339]} & 0.5684 {\scriptsize [0.5593, 0.5773]} \\
  
  MUSE+ 
  & 0.9577 {\scriptsize [0.9546, 0.9608]} & 0.8911 {\scriptsize [0.8798, 0.9012]} & 0.3944 {\scriptsize [0.3575, 0.4306]} & 0.6637 {\scriptsize [0.6458, 0.6809]} & 0.9344 {\scriptsize [0.9312, 0.9377]} & 0.7220 {\scriptsize [0.7050, 0.7380]} & 0.1328 {\scriptsize [0.1200, 0.1491]} & 0.4857 {\scriptsize [0.4849, 0.4866]} & 0.8477 {\scriptsize [0.8426, 0.8530]} & 0.6540 {\scriptsize [0.6426, 0.6653]} & 0.2297 {\scriptsize [0.2176, 0.2429]} & 0.5043 {\scriptsize [0.4965, 0.5117]} \\
  
  AdaCoMed 
  & 0.9363 {\scriptsize [0.9326, 0.9399]} & 0.9334 {\scriptsize [0.9246, 0.9417]} & 0.5481 {\scriptsize [0.5091, 0.5852]} & 0.7306 {\scriptsize [0.7168, 0.7434]} & 0.9412 {\scriptsize [0.9312, 0.9478]} & 0.7522 {\scriptsize [0.7361, 0.7669]} & \underline{0.1512 {\scriptsize [0.1361, 0.1694]}} & 0.4906 {\scriptsize [0.4867, 0.4958]} & 0.8459 {\scriptsize [0.8411, 0.8514]} & 0.6761 {\scriptsize [0.6624, 0.6890]} & 0.1835 {\scriptsize [0.1702, 0.1984]} & 0.5439 {\scriptsize [0.5394, 0.5493]} \\
  
  CRL-MMNAR 
  & \underline{0.9631 {\scriptsize [0.9607, 0.9654]}} & \underline{0.9481 {\scriptsize [0.9398, 0.9565]}} & \textbf{0.6357 {\scriptsize [0.6036, 0.6665]}} & \textbf{0.8085 {\scriptsize [0.7918, 0.8245]}} & 0.9434 {\scriptsize [0.9400, 0.9468]} & 0.6124 {\scriptsize [0.5955, 0.6306]} & 0.0898 {\scriptsize [0.0795, 0.1034]} & 0.5007 {\scriptsize [0.4935, 0.5090]} & \underline{0.8702 {\scriptsize [0.8655, 0.8749]}} & \underline{0.6890 {\scriptsize [0.6746, 0.7033]}} & \textbf{0.4752 {\scriptsize [0.4564, 0.4947]}} & \textbf{0.6180 {\scriptsize [0.6069, 0.6289]}} \\
  
  \midrule 
  
  \modelname{} 
  & \textbf{0.9675 {\scriptsize [0.9648, 0.9700]}} & \textbf{0.9490 {\scriptsize [0.9385, 0.9596]}} & 0.6127 {\scriptsize [0.5791, 0.6462]} & \underline{0.7898 {\scriptsize [0.7839, 0.7958]}} & \textbf{0.9443 {\scriptsize [0.9411, 0.9476]}} & \textbf{0.7971 {\scriptsize [0.7840, 0.8095]}} & \textbf{0.1879 {\scriptsize [0.1693, 0.2105]}} & 0.4857 {\scriptsize [0.4848, 0.4866]} & \textbf{0.8716 {\scriptsize [0.8666, 0.8765]}} & \textbf{0.7584 {\scriptsize [0.7480, 0.7688]}} & \underline{0.4258 {\scriptsize [0.4063, 0.4462]}} & \underline{0.6099 {\scriptsize [0.5993, 0.6203]}} \\
  \bottomrule
\end{tabular}
}

\vspace{0.4cm}
\textbf{(c) MIMIC-III}

\vspace{0.1cm}
\resizebox{\linewidth}{!}{
\begin{tabular}{l *{4}{c} *{4}{c} *{4}{c}}
  \toprule
  \multirow{2}{*}{\textbf{Method}} & \multicolumn{4}{c}{\textbf{Mortality}} & \multicolumn{4}{c}{\textbf{Long LOS}} & \multicolumn{4}{c}{\textbf{Readmission}} \\
  \cmidrule(lr){2-5} \cmidrule(lr){6-9} \cmidrule(lr){10-13}
  & \textbf{ACC} & \textbf{AUROC} & \textbf{AUPRC} & \textbf{mF1} 
  & \textbf{ACC} & \textbf{AUROC} & \textbf{AUPRC} & \textbf{mF1} 
  & \textbf{ACC} & \textbf{AUROC} & \textbf{AUPRC} & \textbf{mF1} \\
  \midrule
  
  Large-Ensemble 
  & 0.9326 {\scriptsize [0.9263, 0.9385]} & 0.9620 {\scriptsize [0.9556, 0.9677]} & 0.7771 {\scriptsize [0.7458, 0.8073]} & 0.8148 {\scriptsize [0.7994, 0.8302]} & 0.6440 {\scriptsize [0.6338, 0.6544]} & 0.7182 {\scriptsize [0.7086, 0.7281]} & 0.7874 {\scriptsize [0.7785, 0.7957]} & \underline{0.7321 {\scriptsize [0.7220, 0.7430]}} & 0.8122 {\scriptsize [0.8009, 0.8237]} & 0.8187 {\scriptsize [0.8035, 0.8338]} & 0.5689 {\scriptsize [0.5360, 0.6003]} & 0.4004 {\scriptsize [0.3891, 0.4114]} \\
  
  MultimodalPred 
  & 0.8940 {\scriptsize [0.8868, 0.9012]} & 0.9515 {\scriptsize [0.9439, 0.9585]} & 0.7017 {\scriptsize [0.6632, 0.7387]} & 0.7590 {\scriptsize [0.7435, 0.7739]} & 0.6335 {\scriptsize [0.6231, 0.6444]} & 0.6981 {\scriptsize [0.6880, 0.7086]} & 0.7755 {\scriptsize [0.7663, 0.7844]} & 0.7160 {\scriptsize [0.7051, 0.7269]} & 0.8287 {\scriptsize [0.8178, 0.8393]} & 0.8018 {\scriptsize [0.7860, 0.8171]} & 0.4945 {\scriptsize [0.4607, 0.5280]} & 0.3270 {\scriptsize [0.3163, 0.3373]} \\
  
  MM-STGNN 
  & 0.9390 {\scriptsize [0.9291, 0.9479]} & 0.9359 {\scriptsize [0.9175, 0.9515]} & 0.7054 {\scriptsize [0.6306, 0.7697]} & 0.8081 {\scriptsize [0.7735, 0.8380]} & 0.6576 {\scriptsize [0.6388, 0.6768]} & 0.7139 {\scriptsize [0.6938, 0.7355]} & \underline{0.8027 {\scriptsize [0.7808, 0.8240]}} & 0.5047 {\scriptsize [0.4844, 0.5255]} & 0.8662 {\scriptsize [0.8514, 0.8800]} & 0.7671 {\scriptsize [0.7311, 0.8001]} & 0.4997 {\scriptsize [0.4385, 0.5548]} & 0.7059 {\scriptsize [0.6771, 0.7335]} \\
  
  M3care 
  & 0.9494 {\scriptsize [0.9453, 0.9532]} & 0.9676 {\scriptsize [0.9644, 0.9705]} & \underline{0.9073 {\scriptsize [0.8870, 0.9251]}} & \underline{0.9080 {\scriptsize [0.8952, 0.9189]}} & 0.6468 {\scriptsize [0.6366, 0.6574]} & 0.7245 {\scriptsize [0.7148, 0.7341]} & 0.7951 {\scriptsize [0.7874, 0.8028]} & 0.7318 {\scriptsize [0.7212, 0.7429]} & 0.7310 {\scriptsize [0.7206, 0.7418]} & \underline{0.8242 {\scriptsize [0.8079, 0.8399]}} & 0.6319 {\scriptsize [0.6029, 0.6604]} & 0.6361 {\scriptsize [0.6235, 0.6484]} \\
  
  MUSE+ 
  & \underline{0.9524 {\scriptsize [0.9478, 0.9569]}} & 0.9710 {\scriptsize [0.9664, 0.9753]} & 0.8755 {\scriptsize [0.8510, 0.8984]} & 0.8880 {\scriptsize [0.8747, 0.9008]} & 0.6613 {\scriptsize [0.6515, 0.6713]} & 0.7325 {\scriptsize [0.7229, 0.7421]} & 0.7981 {\scriptsize [0.7900, 0.8060]} & \textbf{0.7457 {\scriptsize [0.7353, 0.7567]}} & 0.8887 {\scriptsize [0.8813, 0.8962]} & 0.8238 {\scriptsize [0.8084, 0.8398]} & 0.6139 {\scriptsize [0.5843, 0.6438]} & 0.7423 {\scriptsize [0.7261, 0.7580]} \\
  
  AdaCoMed 
  & 0.8734 {\scriptsize [0.8654, 0.8810]} & 0.9472 {\scriptsize [0.9396, 0.9543]} & 0.6651 {\scriptsize [0.6222, 0.7050]} & 0.7356 {\scriptsize [0.7207, 0.7499]} & 0.6962 {\scriptsize [0.6853, 0.7072]} & 0.7127 {\scriptsize [0.7028, 0.7230]} & 0.7839 {\scriptsize [0.7750, 0.7925]} & 0.6939 {\scriptsize [0.6828, 0.7049]} & \underline{0.8919 {\scriptsize [0.8813, 0.9024]}} & 0.8158 {\scriptsize [0.8001, 0.8314]} & 0.5407 {\scriptsize [0.5061, 0.5737]} & 0.2918 {\scriptsize [0.2813, 0.3023]} \\
  
  CRL-MMNAR 
  & 0.9509 {\scriptsize [0.9475, 0.9541]} & \underline{0.9752 {\scriptsize [0.9652, 0.9844]}} & \textbf{0.9294 {\scriptsize [0.9100, 0.9470]}} & \textbf{0.9357 {\scriptsize [0.9246, 0.9463]}} & \underline{0.7251 {\scriptsize [0.7191, 0.7315]}} & \underline{0.7751 {\scriptsize [0.7720, 0.7783]}} & 0.7855 {\scriptsize [0.7833, 0.7875]} & 0.7190 {\scriptsize [0.7124, 0.7257]} & 0.8534 {\scriptsize [0.8482, 0.8581]} & 0.8170 {\scriptsize [0.8036, 0.8294]} & \underline{0.6368 {\scriptsize [0.6150, 0.6562]}} & \textbf{0.7976 {\scriptsize [0.7861, 0.8079]}} \\
  
  \midrule 
  
  \modelname{} 
  & \textbf{0.9526 {\scriptsize [0.9475, 0.9576]}} & \textbf{0.9770 {\scriptsize [0.9723, 0.9812]}} & 0.8392 {\scriptsize [0.8131, 0.8634]} & 0.8642 {\scriptsize [0.8498, 0.8778]} & \textbf{0.7334 {\scriptsize [0.7226, 0.7441]}} & \textbf{0.7864 {\scriptsize [0.7751, 0.7971]}} & \textbf{0.8617 {\scriptsize [0.8508, 0.8718]}} & 0.6983 {\scriptsize [0.6862, 0.7094]} & \textbf{0.8937 {\scriptsize [0.8860, 0.9007]}} & \textbf{0.8306 {\scriptsize [0.8148, 0.8462]}} & \textbf{0.6396 {\scriptsize [0.6106, 0.6679]}} & \underline{0.7644 {\scriptsize [0.7494, 0.7788]}} \\
  \bottomrule
\end{tabular}
}
\end{sidewaystable*}

%% file: table/modality_exp.tex
\begin{table*}[t]
\centering
\caption{
Modality ablation across MIMIC-IV and MIMIC-III.
C, E, R, and D denote chest x-rays, structured EHR events,
radiology reports, and discharge summaries, respectively.
Both MIMIC-IV and MIMIC-III include unimodal and multimodal configurations.
Discharge-summary configurations are not applicable to Long LOS because
the task uses only data recorded during the first 48 hours. Best results are boldfaced separately within each dataset and task.
}
\label{tab:merged_ablation_study}

\small
\setlength{\tabcolsep}{5.5pt}
\renewcommand{\arraystretch}{1.08}

\resizebox{\linewidth}{!}{
\begin{tabular}{l *{4}{c} *{4}{c} *{4}{c}}
\toprule
\multirow{2}{*}{\textbf{Dataset / Input Configuration}}
&
\multicolumn{4}{c}{\textbf{Mortality}}
&
\multicolumn{4}{c}{\textbf{Long LOS}}
&
\multicolumn{4}{c}{\textbf{Readmission}}
\\
\cmidrule(lr){2-5}
\cmidrule(lr){6-9}
\cmidrule(lr){10-13}

&
\textbf{ACC} & \textbf{AUROC} & \textbf{AUPRC} & \textbf{mF1}
&
\textbf{ACC} & \textbf{AUROC} & \textbf{AUPRC} & \textbf{mF1}
&
\textbf{ACC} & \textbf{AUROC} & \textbf{AUPRC} & \textbf{mF1}
\\
\midrule

\rowcolor{mimicfour}
\multicolumn{13}{l}{\textbf{\textit{MIMIC-IV: Unimodal}}}
\\

\rowcolor{mimicfour}
C
& 0.9099 & 0.8734 & 0.5536 & 0.7544
& 0.6489 & 0.6831 & 0.7602 & 0.5085
& 0.8179 & 0.7667 & 0.4736 & 0.6911
\\

\rowcolor{mimicfour}
E
& 0.9542 & 0.8742 & 0.7698 & 0.7871
& 0.6421 & 0.6531 & 0.8105 & 0.5489
& 0.8696 & 0.8109 & 0.6414 & 0.7166
\\

\rowcolor{mimicfour}
R
& 0.8681 & 0.8009 & 0.3920 & 0.6541
& 0.7164 & 0.6555 & 0.8209 & 0.4174
& 0.8225 & 0.7116 & 0.4513 & 0.6001
\\

\rowcolor{mimicfour}
D
& 0.8964 & 0.6947 & 0.1758 & 0.4727
& -- & -- & -- & --
& 0.8048 & 0.6009 & 0.2473 & 0.4459
\\

\addlinespace[2pt]

\rowcolor{mimicthree}
\multicolumn{13}{l}{\textbf{\textit{MIMIC-III: Unimodal}}}
\\

\rowcolor{mimicthree}
E
& 0.9438 & 0.9729 & 0.8240 & 0.8462
& 0.6562 & 0.7626 & 0.8432 & 0.4645
& 0.8831 & 0.8199 & 0.6133 & 0.7496
\\

\rowcolor{mimicthree}
R
& 0.8824 & 0.9010 & 0.5396 & 0.7107
& 0.6569 & 0.7695 & 0.8478 & 0.4710
& 0.8491 & 0.7764 & 0.4489 & 0.6665
\\

\rowcolor{mimicthree}
D
& 0.8619 & 0.8642 & 0.4173 & 0.6787
& -- & -- & -- & --
& 0.8453 & 0.7566 & 0.3960 & 0.6420
\\

\addlinespace[2pt]
\midrule

\rowcolor{mimicfour}
\multicolumn{13}{l}{\textbf{\textit{MIMIC-IV: Multimodal}}}
\\

\rowcolor{mimicfour}
C+E
& 0.8888 & 0.9192 & 0.7035 & 0.7646
& 0.6916 & 0.7001 & 0.7568 & 0.6463
& 0.8775 & \textbf{0.8354} & \textbf{0.6938} & 0.7788
\\

\rowcolor{mimicfour}
C+E+R
& 0.9424 & 0.9570 & 0.8162 & 0.8403
& 0.6906 & 0.7039 & 0.7660 & 0.6402
& 0.8696 & 0.8348 & 0.6887 & 0.7706
\\

\rowcolor{mimicfour}
C+E+D
& 0.9513 & 0.9668 & 0.8538 & 0.8816
& -- & -- & -- & --
& \textbf{0.8848} & 0.8335 & 0.6914 & \textbf{0.7802}
\\

\rowcolor{mimicfour}
E+R
& \textbf{0.9592} & 0.9729 & 0.8743 & \textbf{0.8940}
& 0.6372 & 0.6676 & 0.8181 & 0.5098
& 0.8844 & 0.8237 & 0.6798 & 0.7618
\\

\rowcolor{mimicfour}
E+D
& 0.9557 & 0.9733 & \textbf{0.8747} & 0.8888
& -- & -- & -- & --
& 0.8804 & 0.8225 & 0.6804 & 0.7621
\\

\rowcolor{mimicfour}
E+R+D
& 0.9444 & 0.9706 & 0.8610 & 0.8709
& -- & -- & -- & --
& 0.8824 & 0.8298 & 0.6800 & 0.7636
\\

\addlinespace[2pt]

\rowcolor{mimicthree}
\multicolumn{13}{l}{\textbf{\textit{MIMIC-III: Multimodal}}}
\\

\rowcolor{mimicthree}
E+R
& 0.9313 & 0.9745 & 0.8273 & 0.8283
& 0.6637 & 0.7753 & 0.8503 & 0.4883
& 0.8888 & 0.8301 & 0.6273 & 0.7573
\\

\rowcolor{mimicthree}
E+D
& 0.9463 & 0.9740 & 0.8310 & 0.8492
& -- & -- & -- & --
& \textbf{0.8960} & 0.8240 & 0.6214 & 0.7610
\\

\addlinespace[2pt]
\specialrule{1.2pt}{1pt}{1pt}

\rowcolor{mimicfour}
\textbf{\modelname{} (MIMIC-IV)}
& 0.9528 & \textbf{0.9804} & 0.8605 & 0.8886
& \textbf{0.7409} & \textbf{0.7675} & \textbf{0.8428} & \textbf{0.6750}
& 0.8804 & \textbf{0.8371} & 0.6804 & 0.7567
\\

\rowcolor{mimicthree}
\textbf{\modelname{} (MIMIC-III)}
& \textbf{0.9526} & \textbf{0.9770} & \textbf{0.8392} & \textbf{0.8642}
& \textbf{0.7334} & \textbf{0.7864} & \textbf{0.8617} & \textbf{0.6983}
& 0.8937 & \textbf{0.8306} & \textbf{0.6396} & \textbf{0.7644}
\\

\bottomrule
\end{tabular}
}

\vspace{2pt}
\raggedright
\footnotesize
\textit{Note:} ``--'' indicates that the configuration is not applicable
because terminal discharge summaries are excluded from the first-48-hour
Long LOS setting.
\end{table*}

%% file: table/mask_ablation.tex
\begin{table}
\centering
\caption{Ablation of the TGL attention mask against
All-Causal and Full-Attention alternatives on MIMIC-IV.}
\label{tab:context_ablation}

\setlength{\tabcolsep}{6pt}
\renewcommand{\arraystretch}{1.2}
\resizebox{\linewidth}{!}{
\begin{tabular}{ll ccc}
  \toprule
  \textbf{Task} & \textbf{Metric} & \textbf{TGL} & \textbf{All-Causal} & \textbf{Full-Attention} \\
  \midrule
  \multirow{4}{*}{Mortality}
  & ACC   & 0.9528 & \textbf{0.9533} & 0.9369 \\
  & AUROC & \textbf{0.9804} & 0.9778 & 0.9576 \\
  & AUPRC & \textbf{0.8605} & 0.8500 & 0.7911 \\
  & mF1   & \textbf{0.8886} & 0.8864 & 0.8275 \\
  \midrule
  \multirow{4}{*}{Long LOS}
  & ACC   & \textbf{0.7409} & 0.7305 & 0.7193 \\
  & AUROC & \textbf{0.7675} & 0.7583 & 0.7462 \\
  & AUPRC & \textbf{0.8428} & 0.8346 & 0.8102 \\
  & mF1   & 0.6750 & \textbf{0.6967} & 0.6752 \\
  \midrule
  \multirow{4}{*}{Readmission}
  & ACC   & 0.8804 & \textbf{0.8819} & 0.8651 \\
  & AUROC & \textbf{0.8371} & 0.8271 & 0.8042 \\
  & AUPRC & \textbf{0.6804} & 0.6804 & 0.6323 \\
  & mF1   & \textbf{0.7567} & 0.7535 & 0.7348 \\
  \bottomrule
\end{tabular}
}
\end{table}

%% file: table/contrast_bce.tex
\begin{table*}[t]
\centering
\caption{Phenotype-stratified performance of the $L_{\mathrm{pred}}$-only
baseline and the change obtained by adding $L_{\mathrm{pheno}}$ across
datasets. $\Delta_{\mathrm{pheno}}$ denotes the performance under the
joint objective minus that under the prediction-only objective; positive
improvements are boldfaced. Avg. denotes the unweighted mean across
phenotypes.}
\label{tab:pcl_bce}

\setlength{\tabcolsep}{2.6pt}
\renewcommand{\arraystretch}{1.02}
\scriptsize
\resizebox{0.9\textwidth}{!}{
\begin{tabular}{c l r *{12}{c}}
\toprule

\multirow{3}{*}{\textbf{Dataset}}
&
\multirow{3}{*}{\textbf{Phenotype}}
&
\multirow{3}{*}{\textbf{$n$}}
&
\multicolumn{4}{c}{\textbf{Mortality}}
&
\multicolumn{4}{c}{\textbf{Long LOS}}
&
\multicolumn{4}{c}{\textbf{Readmission}}
\\

\cmidrule(lr){4-7}
\cmidrule(lr){8-11}
\cmidrule(lr){12-15}

&
&
&
\multicolumn{2}{c}{AUROC}
&
\multicolumn{2}{c}{AUPRC}
&
\multicolumn{2}{c}{AUROC}
&
\multicolumn{2}{c}{AUPRC}
&
\multicolumn{2}{c}{AUROC}
&
\multicolumn{2}{c}{AUPRC}
\\

\cmidrule(lr){4-5}
\cmidrule(lr){6-7}
\cmidrule(lr){8-9}
\cmidrule(lr){10-11}
\cmidrule(lr){12-13}
\cmidrule(lr){14-15}

&
&
&
$L_{\mathrm{pred}}$ & $\Delta_{\mathrm{pheno}}$
&
$L_{\mathrm{pred}}$ & $\Delta_{\mathrm{pheno}}$
&
$L_{\mathrm{pred}}$ & $\Delta_{\mathrm{pheno}}$
&
$L_{\mathrm{pred}}$ & $\Delta_{\mathrm{pheno}}$
&
$L_{\mathrm{pred}}$ & $\Delta_{\mathrm{pheno}}$
&
$L_{\mathrm{pred}}$ & $\Delta_{\mathrm{pheno}}$
\\

\midrule

\multirow{7}{*}{\rotatebox[origin=c]{90}{\textit{MIMIC-IV}}}

& Circulatory
& 764
& .9708 & \gain{.0061}
& .8272 & \gain{.0131}
& .7899 & \gain{.0042}
& .8331 & \gain{.0048}
& .8164 & \dropval{.0015}
& .6596 & \dropval{.0044}
\\

& Trauma \& Poisoning
& 112
& .9374 & \gain{.0120}
& .7273 & \gain{.0821}
& .7490 & \gain{.0127}
& .7948 & \gain{.0143}
& .8920 & \gain{.0088}
& .6345 & \dropval{.0070}
\\

& Endocrine \& Metabolic
& 206
& .9805 & \gain{.0073}
& .8428 & \gain{.0688}
& .7517 & \gain{.0189}
& .8686 & \gain{.0130}
& .7788 & \gain{.0040}
& .6280 & \dropval{.0113}
\\

& Respiratory
& 152
& .9582 & \gain{.0117}
& .8273 & \gain{.0517}
& .7492 & \gain{.0254}
& .8570 & \gain{.0239}
& .7868 & \gain{.0047}
& .6901 & \dropval{.0170}
\\

& Digestive
& 127
& .9860 & \gain{.0013}
& .9255 & \dropval{.0062}
& .9193 & \gain{.0111}
& .9863 & \gain{.0024}
& .8407 & \gain{.0043}
& .7428 & \dropval{.0012}
\\

& Neoplasms
& 64
& .9259 & \gain{.0189}
& .6122 & \gain{.0706}
& .7954 & \gain{.0103}
& .7429 & \dropval{.0011}
& .7129 & \dropval{.0049}
& .4387 & \gain{.0079}
\\

\cmidrule(lr){2-15}

& \textbf{Avg.}
& \textbf{1425}
& .9598 & \gain{.0096}
& .7937 & \gain{.0467}
& .7924 & \gain{.0138}
& .8471 & \gain{.0096}
& .8046 & \gain{.0026}
& .6323 & \dropval{.0055}
\\

\midrule

\multirow{7}{*}{\rotatebox[origin=c]{90}{\textit{eICU}}}

& Circulatory
& 7205
& .9517 & \gain{.0102}
& .6472 & \gain{.0622}
& .7785 & \gain{.0142}
& .1541 & \gain{.0011}
& .7819 & \dropval{.0008}
& .4947 & \dropval{.0098}
\\

& Trauma \& Poisoning
& 867
& .9499 & \gain{.0291}
& .6337 & \gain{.1476}
& .7269 & \gain{.0152}
& .1443 & \gain{.0070}
& .8331 & \dropval{.0007}
& .4111 & \gain{.0350}
\\

& Endocrine \& Metabolic
& 733
& .9255 & \gain{.0060}
& .2681 & \gain{.0223}
& .7122 & \gain{.0709}
& .1032 & \gain{.0493}
& .7577 & \gain{.0120}
& .1675 & \gain{.0199}
\\

& Respiratory
& 2507
& .8735 & \gain{.0521}
& .4697 & \gain{.0417}
& .7169 & \gain{.0209}
& .1634 & \gain{.0178}
& .7243 & \gain{.0076}
& .4056 & \dropval{.0017}
\\

& Digestive
& 1575
& .9450 & \gain{.0056}
& .6688 & \gain{.0153}
& .8203 & \gain{.0175}
& .2117 & \gain{.0283}
& .7598 & \gain{.0105}
& .4708 & \gain{.0132}
\\

& Neoplasms
& 193
& .8372 & \gain{.0614}
& .3181 & \gain{.0946}
& .7351 & \dropval{.0405}
& .4892 & \dropval{.1602}
& .7241 & \gain{.0162}
& .3546 & \gain{.0173}
\\

\cmidrule(lr){2-15}

& \textbf{Avg.} & \textbf{13080}
& .9138 & \gain{.0274}
& .5009 & \gain{.0640}
& .7483 & \gain{.0164}
& .2110 & \dropval{.0095}
& .7635 & \gain{.0075}
& .3841 & \gain{.0123}

\\

\midrule

\multirow{7}{*}{\rotatebox[origin=c]{90}{\textit{MIMIC-III}}}

& Circulatory
& 3363
& .9816 & \gain{.0011}
& .8382 & \gain{.0024}
& .7676 & \gain{.0031}
& .8464 & \gain{.0020}
& .8169 & \gain{.0059}
& .6073 & \gain{.0009}
\\

& Trauma \& Poisoning
& 647
& .9788 & \dropval{.0012}
& .8393 & \dropval{.0116}
& .8291 & \gain{.0065}
& .8825 & \gain{.0081}
& .8833 & \dropval{.0007}
& .6925 & \dropval{.0189}
\\

& Endocrine \& Metabolic
& 553
& .9818 & \gain{.0007}
& .8258 & \gain{.0001}
& .7773 & \gain{.0056}
& .8171 & \gain{.0037}
& .8186 & \gain{.0182}
& .6252 & \dropval{.0117}
\\

& Respiratory
& 548
& .9517 & \dropval{.0007}
& .7846 & \gain{.0146}
& .7576 & \dropval{.0096}
& .8840 & \dropval{.0025}
& .8297 & \gain{.0107}
& .6584 & \gain{.0156}
\\

& Digestive
& 547
& .9739 & \gain{.0026}
& .9052 & \gain{.0012}
& .8115 & \dropval{.0014}
& .9028 & \dropval{.0032}
& .8289 & \dropval{.0005}
& .7245 & \gain{.0046}
\\

& Neoplasms
& 231
& .9690 & \gain{.0133}
& .7782 & \gain{.0564}
& .7621 & \gain{.0040}
& .8422 & \gain{.0053}
& .8157 & \gain{.0412}
& .6091 & \gain{.0222}
\\

\cmidrule(lr){2-15}

& \textbf{Avg.}
& \textbf{5889}
& .9728 & \gain{.0026}
& .8286 & \gain{.0105}
& .7842 & \gain{.0014}
& .8625 & \gain{.0022}
& .8322 & \gain{.0124}
& .6528 & \gain{.0021}
\\

\bottomrule
\end{tabular}
}
\end{table*}

%% file: discussion.tex
This study examined whether admission-level risk prediction benefits from
representing heterogeneous clinical data as a temporally organized trajectory.
Across datasets and tasks, the results indicate that temporal organization
provides a complementary modeling dimension beyond multimodal fusion and
missing-modality handling. The case analyses in
Section~\ref{sec:trajectory} further illustrate this distinction:
admission-level context contributed strongly to early Long LOS estimates,
whereas mortality and readmission estimates continued to change as new
day-specific findings were recorded.

The TGL attention ablation supports the proposed global--local structure. Compared with All-Causal and Full-Attention alternatives, TGL achieved the highest AUROC across all three tasks. All-Causal forces global context through the same strict temporal ordering as local events, delaying its influence on early predictions such as Long LOS; Full-Attention removes this ordering entirely, allowing later local events to inform earlier-day representations in a way that does not reflect how the record is actually observed. TGL's intermediate structure -- global context always accessible, local events temporally ordered -- appears to better match how admission-level and day-specific information should be integrated. The gains were more consistent for AUROC and AUPRC than for mF1, indicating that the main benefit lies in ranking discrimination rather than uniform improvement across threshold-dependent metrics.

Modality contributions were both task- and time-dependent. Structured EHR data
were particularly informative for mortality and readmission, while chest x-rays provided stronger unimodal discrimination for Long LOS on
MIMIC-IV. Multimodal combinations generally improved performance, but the
trajectory analyses showed that the relative influence of each modality shifted
over the course of an admission. These findings support an event-driven
representation in which available inputs are incorporated according to when
they are recorded, rather than assigned a fixed importance across tasks or
patients.

Phenotype-guided contrastive learning provided an additional but less uniform
benefit. Defining positive pairs by both outcome status and diagnosis-derived
phenotype improved mean phenotype-stratified AUROC across all nine
dataset--task settings, although AUPRC and individual phenotype effects varied.
This supports its role as a phenotype-aware representation regularizer,
while also motivating further analysis of phenotype-specific effects.

\section{Limitations and Future Work}
This study has several limitations. 
First, \modelname{} operates at hospital-day resolution, which reduces sparsity and sequence length but may miss rapidly evolving intra-day physiology or treatment response. Our case studies (Figure~\ref{fig:trajectory_cases}) illustrate this: responses were generally prompt for same-day evidence but muted when new findings appeared in a modality with low attribution at that stage. Finer temporal resolutions may be more appropriate for acute outcomes.
Second, baseline comparisons should be read alongside our controlled modality, attention, and objective ablations, given differences in architecture and temporal organization. Finally, evaluation was limited to public retrospective datasets with varying modality configurations; future work should examine finer temporal resolutions, alternative phenotypes, prospective and external validation, and deployment settings with evolving observation windows and modality availability.

%% file: 4.conclusion.tex
We introduced \modelname{}, a daily-scale multimodal framework that represents
heterogeneous inpatient data as a temporally organized global--local
trajectory. Across MIMIC-IV, eICU, and MIMIC-III, \modelname{} achieved the
highest AUROC in all nine dataset--task settings for in-hospital mortality,
early Long LOS, and 30-day readmission prediction. The results show that
temporal organization is a complementary design dimension to multimodal fusion
and missing-modality handling, while phenotype-guided contrastive learning
provides additional representation-level benefits.